\documentclass[10pt,twocolumn,letterpaper]{article}

\usepackage[pagenumbers]{cvpr} 

\usepackage{makecell} 
\usepackage{algorithm}
\usepackage{algpseudocode}
\usepackage{multirow}
\usepackage{paralist}
\usepackage{adjustbox}
\usepackage{bbding}

\definecolor{cvprblue}{rgb}{0.21,0.49,0.74}
\usepackage[pagebackref,breaklinks,colorlinks,allcolors=cvprblue]{hyperref}

\def\paperID{1562} 
\def\confName{CVPR}
\def\confYear{2026}

\title{UniRain: Unified Image Deraining with RAG-based Dataset Distillation and Multi-objective Reweighted Optimization}

\author{
Qianfeng Yang$^1$ \quad
Qiyuan Guan$^1$ \quad
Xiang Chen$^2$ \quad
Jiyu Jin$^{1*}$ \quad
Guiyue Jin$^1$ \quad
Jiangxin Dong$^2$\\[0.2em]
$^1$Dalian Polytechnic University \quad
$^2$Nanjing University of Science and Technology\\
{\small\url{https://lowlevelcv.com/}} \quad {\small\url{https://github.com/QianfengY/UniRain}}
}

\begin{document}
\maketitle

\begin{abstract}
Despite significant progress has been made in image deraining, we note that most existing methods are often developed for only specific types of rain degradation and fail to generalize across diverse real-world rainy scenes.
How to effectively model different rain degradations within a universal framework is important for real-world image deraining.
In this paper, we propose UniRain, an effective unified image deraining framework capable of restoring images degraded by rain streak and raindrop under both daytime and nighttime conditions.
To better enhance unified model generalization, we construct an intelligent retrieval augmented generation (RAG)-based dataset distillation pipeline that selects high-quality training samples from all public deraining datasets for better mixed training.
Furthermore, we incorporate a simple yet effective multi-objective reweighted optimization strategy into the asymmetric mixture-of-experts (MoE) architecture to facilitate consistent performance and improve robustness across diverse scenes.
Extensive experiments show that our framework performs favorably against the state-of-the-art models on our proposed benchmarks and multiple public datasets.
%
\end{abstract}

\vspace{-3mm}
\section{Introduction}
\label{sec:intro}
Recent years have witnessed the emergence of diverse rain datasets~\cite{guo2023sky,ba2022not,jiang2020multi} and the development of numerous deep learning–based deraining methods~\cite{song2024learning,chen2024rethinking, jiang2021rain}.
As the field has evolved, researchers have increasingly focused on addressing different rain degradation patterns, such as rain streaks~\cite{yang2017deep}, raindrops~\cite{qian2018attentive}, nighttime rain~\cite{guan2025cstnet}, and so on.
Such clear settings enable researchers to build models adapted to the specific characteristics of each rainy scenario.

\begin{figure}[!t]
	\centering
	\begin{subfigure}{0.97\linewidth}
        \centering
        \includegraphics[width=\linewidth]{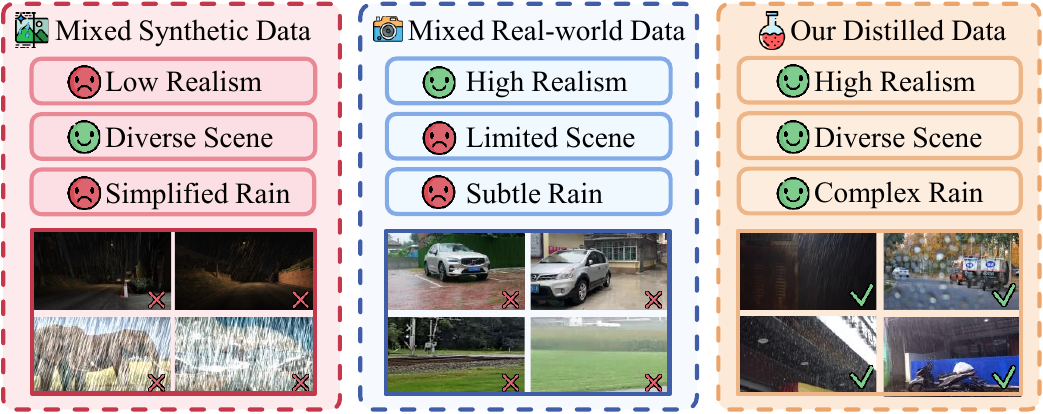}
        \caption{Visualization of rainy images from different mixed datasets }
        \label{fig:intro_a}
    \end{subfigure} \\
    \begin{subfigure}{0.2\linewidth}
        \centering
        \includegraphics[width=\linewidth]{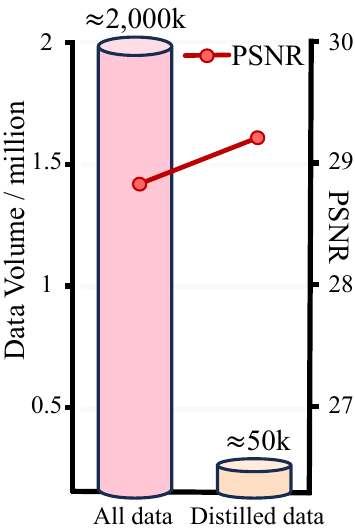}
        \caption{Dataset}
        \label{fig:intro_b}
    \end{subfigure} 
    \begin{subfigure}{0.367\linewidth}
        \centering
        \includegraphics[width=\linewidth]{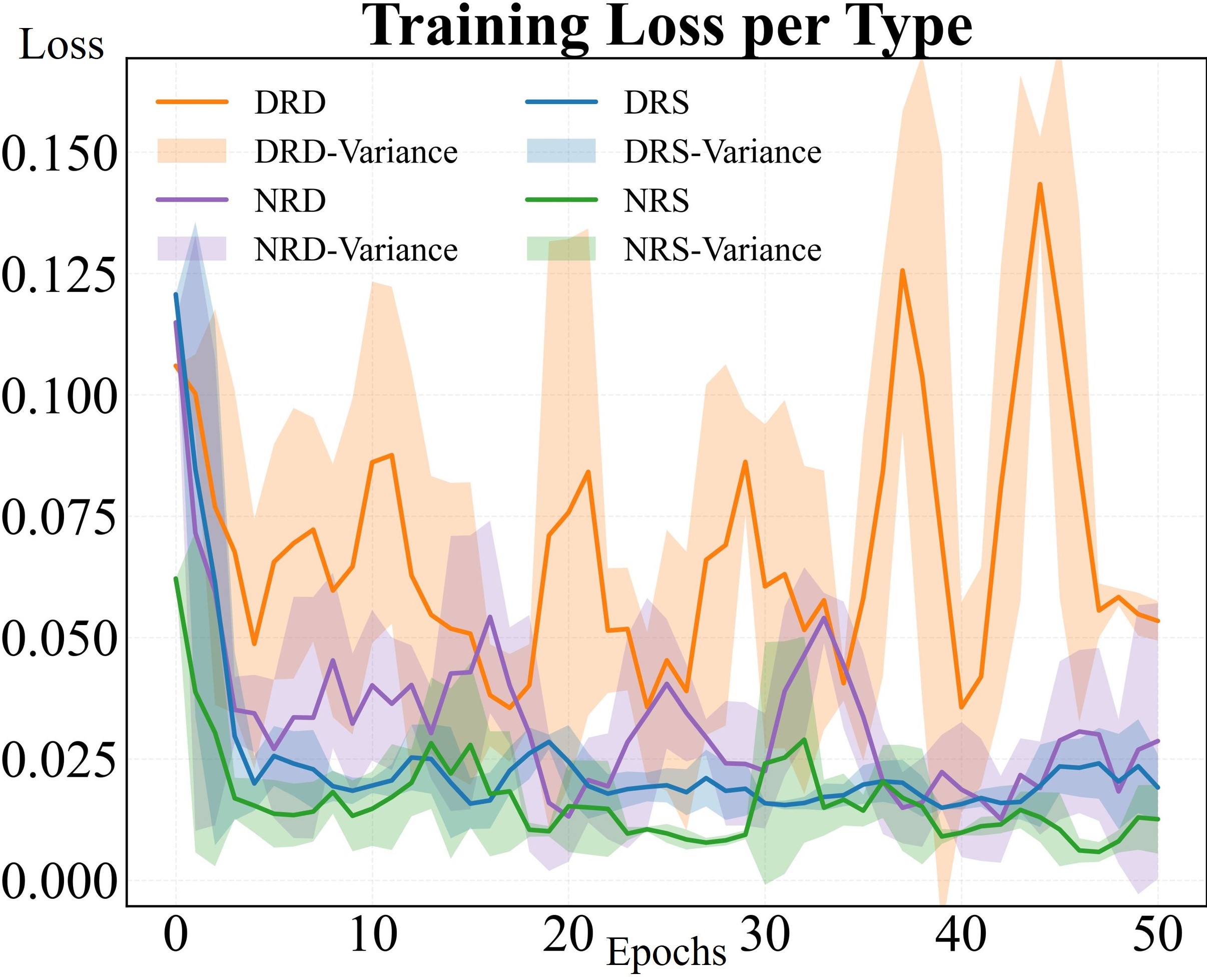}
        \caption{Training imbalance}
        \label{fig:intro_c}
    \end{subfigure}
    \begin{subfigure}{0.375\linewidth}
        \centering
        \includegraphics[width=\linewidth]{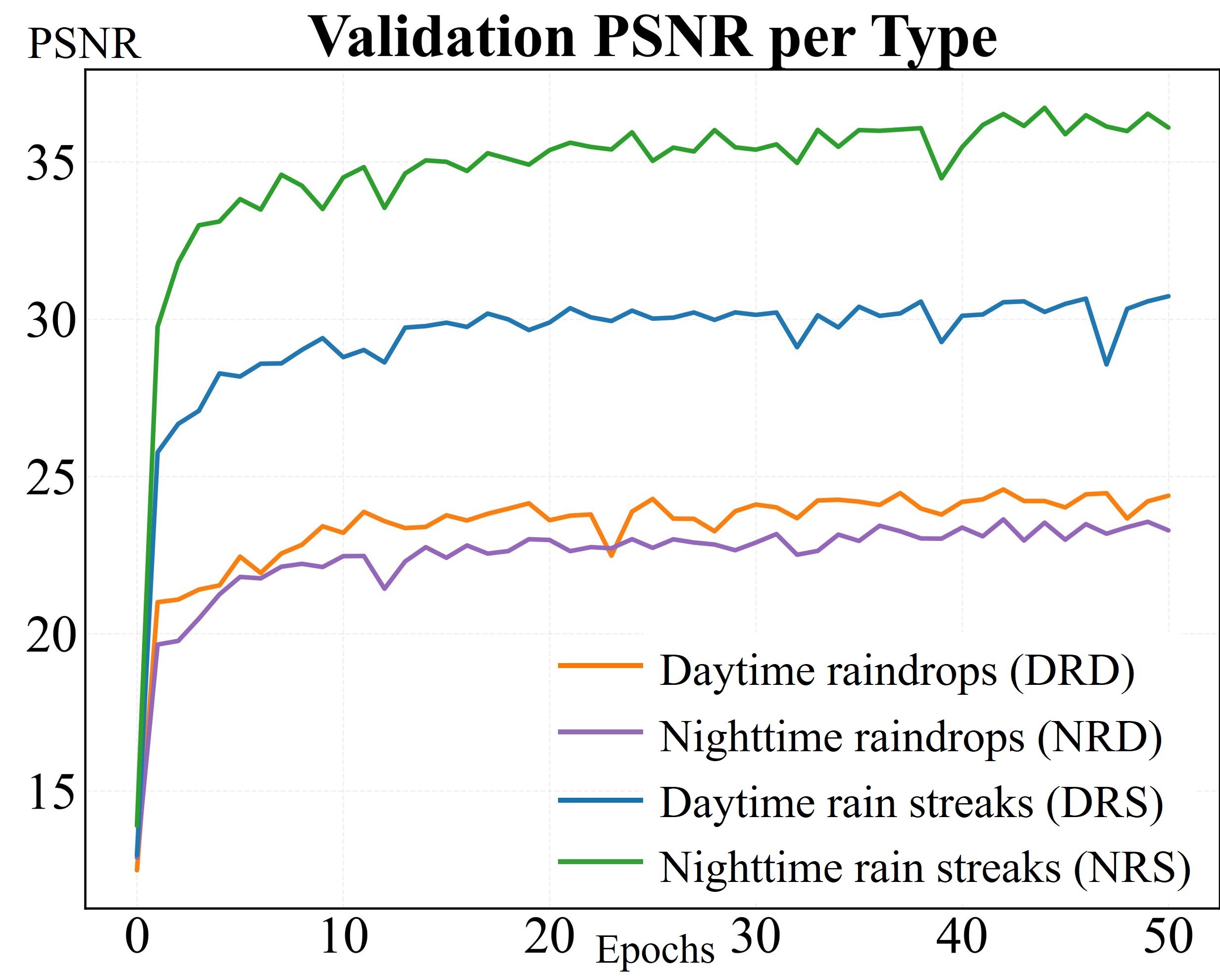}
        \caption{Performance difference}
        \label{fig:intro_d}
    \end{subfigure}
	\vspace{-3mm}
	\caption{Overview of \textbf{motivation}.
    (a) Rainy image samples from public datasets, illustrating the noticeable differences in data quality.
    (b) Directly merging existing synthetic and real datasets enlarges data volume, yet quality disparity hinders performance, as shown by PSNR results.
    (c) The loss curves of DRS, DRD, NRS, and NRD (denoting daytime/nighttime rain streaks and raindrops) show different convergence rates, leading to imbalance in unified training.
    (d) The PSNR curves indicate that the model tends to favor simpler degradations but struggles with complex ones.
    }
	\label{fig:intro}
	\vspace{-4mm}
\end{figure}

However, these models are typically designed for specific rain degradation patterns, and their performance often drops significantly when applied to other types of rain.
Real-world rainy scenes are highly complex and dynamic. For example, daytime rain is mainly characterized by streak degradations, while nighttime scenes, affected by low illumination and varying light sources, amplify the complexity of rain artifacts.
Intelligent systems need to handle multiple rain degradation types, but relying on separate models for different conditions makes switching cumbersome and significantly reduces deployment efficiency~\cite{yan2025towards, KOU2025111172,li2025foundir}.
However, few effort has been made in unified image deraining due to the absence of mixed datasets that integrate multiple rain degradation types.
Thus, it is of great interest to unlock a unified image deraining method that can simultaneously handle multiple rain degradations within a single model.

To achieve better unified image deraining, two key factors should be considered: constructing mixed dataset and designing unified model.
Regarding the \textbf{mixed dataset}, the most direct strategy is to merge all synthetic and real-world rain datasets (\textit{i.e.}, $>$ 2,000k pairs)~\cite{yang2017deep, jiang2020multi, qian2018attentive, quan2021removing, zhang2022gtav}. 
However, we note that these public datasets vary noticeably in quality, as some visual samples illustrated in Figure~\ref{fig:intro_a}.
Such uneven data quality may lead to introduce inaccurate supervisory signals during training, which hinders model convergence~\cite{liang2025enhancing}.
This instability further causes undesirable interference and weakens the model’s ability to generalize across diverse rain conditions, as shown in Figure~\ref{fig:intro_b}.
These insights motivate us to distill reliable data from large-scale public datasets for unified image deraining.

For the \textbf{unified model}, how to equip it with consistent capability across diverse rain conditions is of vital importance.
As depicted in Figure~\ref{fig:intro_c}, we find that different types of rain degradations exhibit varying levels of difficulty and distinct convergence behaviors during training.
If we train mixed rain-type data with the same optimization objective, it would lead to imbalance across different degradation types~\cite{gong2024coba}.
In this case, the model tends to overfit easier rain types (\textit{e.g.}, nighttime rain streaks) while struggling with more complex rain degradations (\textit{e.g.}, daytime raindrops), resulting in uneven restoration quality, as illustrated in Figure~\ref{fig:intro_d}.
This motivates us to introduce a multi-objective optimization strategy to balance the deraining capability of the unified model across different rain types.

To address these issues, we propose UniRain, an effective unified image deraining framework capable of restoring images degraded by rain streak and raindrop under both daytime and nighttime conditions.
Specifically, we first construct an intelligent retrieval augmented generation (RAG)-based dataset distillation pipeline that distills reliable training samples from all public deraining datasets for better mixed training.
We build a database of real-world rainy images to guide vision-language models (VLMs) in assessing data quality, thereby enabling large-scale dataset evaluation and distillation of high-quality samples for unified image deraining.
Furthermore, we develop an asymmetric mixture-of-experts (MoE) architecture, where the encoder integrates the soft-MoE to collaboratively preserve diverse degradation cues, while the decoder employs the hard-MoE with top-k routing to enhance fine texture reconstruction.
To achieve better training balance, we incorporate a simple yet effective multi-objective reweighted optimization strategy into the framework to harmonize the learning of different rain degradation types.
Extensive experiments demonstrate that our framework performs consistently favorably against the state-of-the-art models on both our proposed benchmarks and multiple public datasets.

We summarize our main contributions as follows:

\begin{compactitem}
\item We propose UniRain, an effective unified image deraining model capable of handling rain streaks and raindrops under daytime and nighttime conditions.
	
\item We develop an intelligent RAG-based data distillation pipeline that filters reliable training samples from public datasets to improve training on mixed data.
	
\item We introduce a simple yet effective multi-objective reweighted optimization strategy to harmonize the learning of multiple types and alleviate task imbalance.

\item Extensive experiments demonstrate that our UniRain achieves consistently competitive performance in terms of both quantitative metrics and visual quality.
\end{compactitem}
\begin{figure*}[t]
	\centering
	\includegraphics[width=1.0\textwidth]{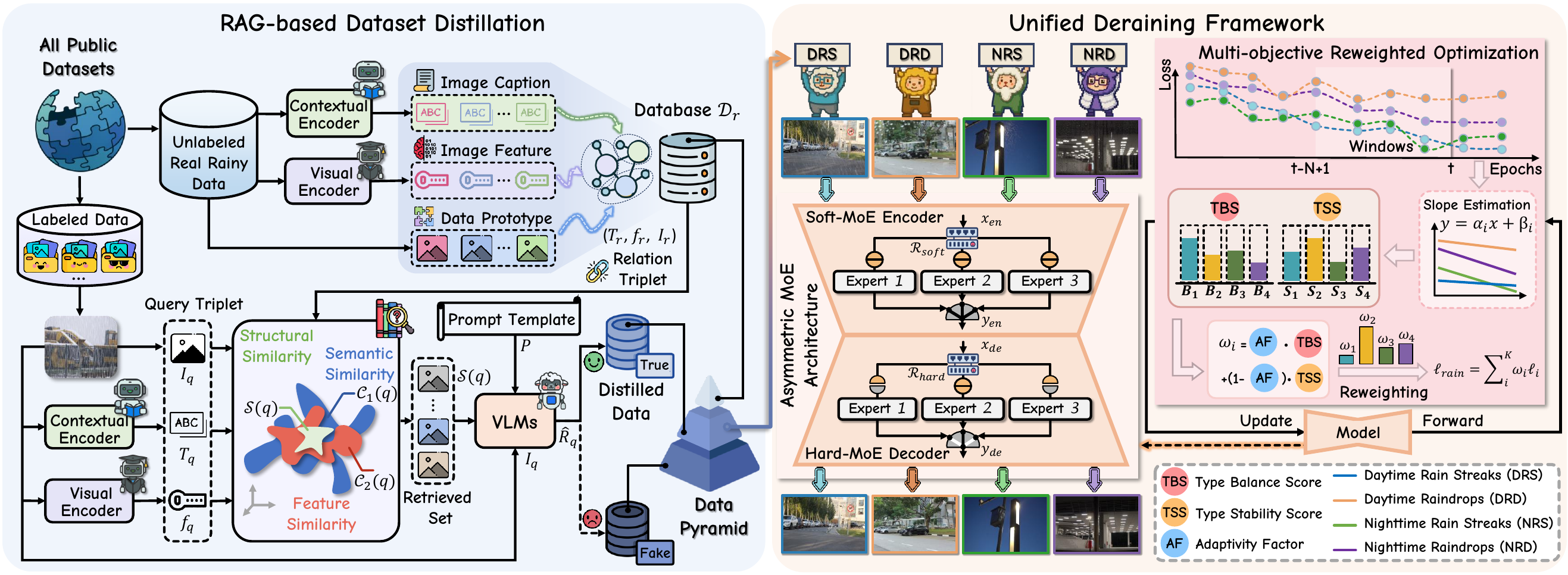}
	\vspace{-6mm}
	\caption{Overall framework of \textbf{UniRain}.~(Left) The RAG-based dataset distillation pipeline retrieves real rainy references consistent with the query image via multi-level similarity search and employs vision language models to evaluate its quality, thereby distilling reliable samples from public datasets.~(Right) The asymmetric MoE architecture consists of soft-MoE encoder and hard-MoE decoder, optimized via the multi-objective reweighted strategy to achieve balanced learning and robust performance across multiple rain degradation types.}
	\label{fig:pipeline}
	\vspace{-3mm}
\end{figure*}

\section{Related Work}
\label{sec:related}

{\flushleft\textbf{Deep image deraining}.}
In recent years, the field of single image deraining has witnessed the emergence of numerous approaches and benchmark datasets~\cite{chen2025towards}.
With respect to \textbf{deraining datasets}, numerous synthetic datasets featuring different types of rain degradation have been proposed, including rain streaks (\textit{e.g.}, Rain200L/H~\cite{yang2017deep}, Rain13k~\cite{jiang2020multi}), raindrops (\textit{e.g.}, RainDrop~\cite{qian2018attentive}, RainDS~\cite{quan2021removing}), and nighttime rain (\textit{e.g.}, GTAV-NightRain~\cite{zhang2022gtav}, HQ-NightRain~\cite{guan2025cstnet}).
To boost real-world deraining performance, several real-world datasets have been proposed, including SPA-Data~\cite{wang2019spatial}, GT-Rain~\cite{ba2022not}, and LHP-Rain~\cite{guo2023sky}.
It is worth noting that significant differences exist among these datasets in terms of background quality, image resolution, and the formation patterns of rain.
In this work, we propose a dataset distillation strategy to select high-quality training samples from all public datasets, aiming to better facilitate unified image deraining.

Regarding the \textbf{deraining models}, we note that most existing approaches remain limited to handling only specific types of rain degradation, without considering a unified solution that generalizes across diverse real-world rainy scenarios. 
Inspired by the concept of all-in-one image restoration, Yan first contributed a universal deraining network developed for multiple rain degradation types in driving scenarios~\cite{yan2025towards}.
To break through the generalization bottleneck of image deraining, our goal is to develop a unified deraining framework capable of simultaneously handling multiple rain degradation types across diverse real-world scenarios.

\vspace{-2mm}

{\flushleft\textbf{Retrieval augmented generation (RAG)}.}
RAG refers to the process of retrieving relevant information from external knowledge bases prior to generating outputs with large language models (LLMs)~\cite{lewis2020retrieval}.
This allows models to leverage domain-specific information and generate more informed outputs without requiring retraining~\cite{gao2023retrieval}.
As a cutting-edge technique, it has been applied in high-level vision tasks, \textit{\textit{e.g.}}, image analysis~\cite{zhang2024imagerag,li2025surgical}, image synthesis~\cite{blattmann2022semi, mukherjee2024rissole}, and 3D modeling~\cite{wang2024phidias,yang2025rethinking}.
Recent studies have begun exploring the potential of RAG in low-level vision tasks.
For example, Guo~\textit{et al.}~\cite{guo2024refir} proposed ReFIR, a training-free retrieval-augmented image restoration framework that retrieves content-relevant high-quality references and injects them into large restoration models to boost fidelity.
Inspired by this popular technique, we first leverage RAG in the process of dataset distillation to better facilitate model training.

\vspace{-2mm}

{\flushleft\textbf{Multi-objective learning}.}
Multi-objective learning jointly optimizes models across multiple tasks, leveraging shared representations to capture common underlying structures among related tasks~\cite{crawshaw2020multi}.
In the context of optimizing multiple objectives that may exhibit conflicting gradients, multi-objective learning has been extensively investigated to improve both optimization stability and overall task performance.
To this end, numerous techniques have been proposed, including multi-armed bandit~\cite{slivkins2019introduction}, gradient manipulation approaches~\cite{yu2020gradient} and self-paced learning~\cite{kumar2010self}.
For low-level vision tasks, Sun~\textit{et al.}~\cite{sun2024perception} proposed a hybrid evolutionary gradient optimizer to balance perceptual quality and reconstruction fidelity through multi-objective optimization.
In this work, we propose a dynamic reweighted optimization mechanism to adaptively balance multiple rain degradation removal objectives for unified image deraining.

\section{Methodology}
Our goal is to formulate an effective unified image deraining model that can address multiple types of rain degradations.
Towards this goal, we develop an intelligent RAG-based dataset distillation pipeline that evaluates large-scale public datasets and selects high-quality samples for more balanced mixed training, as illustrated in Figure~\ref{fig:pipeline} (left).
Furthermore, to achieve consistent performance and improved robustness, we incorporate a simple yet effective multi-objective reweighted optimization strategy into the asymmetric MoE architecture, as shown in Figure~\ref{fig:pipeline} (right).
In what follows, we present the details of the proposed approach.

\label{sec:method}
\subsection{RAG-based dataset distillation}
If all synthetic and real datasets are directly mixed for training, the uneven data quality may introduce interference and consequently degrade the model performance.
To maximize the effectiveness of mixed training dataset, we construct the RAG-based dataset distillation pipeline.
The pipeline mainly consists of retrieval stage, which retrieves relevant real-world reference, and generation stage, which assesses data quality and distills reliable samples for training.

\vspace{-2mm}
{\flushleft\textbf{Retrieval stage}.}
This stage start from a large-scale knowledge corpus of millions of synthetic and real rainy images, partitioned into unlabeled real data for database construction and labeled data for distillation.
For each real image $I_r$, a textual caption $T_r$ is generated using the contextual encoder BLIP~\cite{li2022blip}, and a visual encoder CLIP~\cite{radford2021learning} is used to extract its feature embedding $f_r$. The resulting relation triplet $(T_r,f_r,I_r)$ is then stored in the database $\mathcal{D}_r$.
For each query image, we perform a similar process~\cite{li2022blip, qwen2.5,grootendorst2020keybert} to obtain its relation triplet $(T_q, f_q, I_q)$.
Given the query triplet, we adopt a hierarchical similarity matching process that considers semantic, feature, and structural similarities to retrieve the most relevant samples from $\mathcal{D}_r$.
Specifically, we first compute the semantic similarity score $s_{\text{txt}}(q, r)$:
\begin{eqnarray}
    \begin{aligned}
        s_{txt}(q,r)=\parallel \phi_T(T_q)- \phi_T(T_r) \parallel_2,
    \end{aligned}
\label{eq1}
\end{eqnarray}
where $\phi_T(\cdot)$ denotes the CLIP text encoder and $\parallel \cdot \parallel_2$ denotes the $L_2$ norm. The top $K_1$ samples with the highest semantic similarity scores form the initial candidate set $\mathcal{C}_1(q)$.
Within this candidate set, we further compute the visual feature similarity scores $s_{vis}(q,r^\prime)$ using cosine similarity to assess appearance closeness, as defined below:
\begin{eqnarray}
    \begin{aligned}
        s_{vis}(q,r^\prime)=\frac{f_{q}^{^{\top}}f_{r^\prime}}{\left \| f_{q} \right \|_2 \left \| f_{r^\prime} \right \|_2} , \quad r^\prime\in {\mathcal{C}_{1}(q)}.
    \end{aligned}
\label{eq2}
\end{eqnarray}

The candidate set $\mathcal{C}_2(q)$ is constructed, retaining the top $K_2$ most similar samples.
Finally, we employ the $SSIM(\cdot)$ function to evaluate structural similarity and capture fine-grained perceptual consistency, as defined below:
%
\begin{eqnarray}
    \begin{aligned}
        s_{perc}(q,{r}'')=SSIM(I_q,I_{{r}''}) , \quad  {r}''\in {\mathcal{C}_{2}(q)}.
    \end{aligned}
\label{eq3}
\end{eqnarray}

The top $K_3$ samples are selected based on the structural similarity scores $s_{\text{perc}}(q, r'')$ to construct $\mathcal{S}(q)$.
Through this hierarchical similarity matching process, the pipeline ensures that $\mathcal{S}(q)$ contains the most semantically and visually consistent real rainy scenes, which serve as authentic references for data distillation.

\vspace{-2mm}
{\flushleft\textbf{Generation stage}.}
Once the retrieved set $\mathcal{S}(q)$ is obtained, it is combined with the query image $I_q$ and a predefined prompt template $P$ as input to the VLMs. The overall generation process is expressed as follows:
\begin{eqnarray}
    \begin{aligned}
        R_q = \mathcal{V}(I_q,\mathcal{S}(q),P),
    \end{aligned}
\label{eq4}
\end{eqnarray}
where $\mathcal{V}(\cdot)$ is the VLM, and $R_q$ represents the VLM’s response.
To enhance reliability, we adopt an ensemble voting strategy across three independent VLMs (InternVL2.5-8B~\cite{chen2024expanding}, LLaVA-NeXT-7B~\cite{li2024llava}, and MobileVLM-3B~\cite{wu2024mobilevlm}).
Let $R_q^i \in \{0,1\}$ denote the binary response from the $i$-th model, where 1 indicates True and 0 indicates False.
The final decision is determined by majority voting:
\begin{equation}
\hat{R}_q =
\begin{cases}
1 & \text{if } {\textstyle\sum_{i=1}^{3}}\mathbb{I}(R_q^i=1)\ge 2,\\
0 & \text{otherwise},
\end{cases}
\label{eq5}
\end{equation}
where $\mathbb{I}(\cdot)$ is the indicator function, and $\hat{R}_q=1$ indicates the query rainy image is considered reliable.
Finally, this process forms a data-quality pyramid, where the top layer represents real rainy data, the middle layer consists of distilled high-quality and reliable data close to real distributions, and the bottom layer contains low-fidelity samples. Through this hierarchical refinement, the system distills millions of images into a reliable dataset.

\subsection{Multi-objective reweighted optimization}
Training a single model to jointly handle different rain types introduces optimization imbalance.
The shared optimization objective tends to favor easier rain type with faster loss decay, while more difficult ones remain under-optimized. Consequently, the learned representations become biased, limiting the model’s overall generalization across diverse degradations.
To address the above imbalance, we propose multi-objective reweighted optimization, an adaptive strategy inspired by~\cite{gong2024coba}, which dynamically balances convergence rates and harmonizes multi-objective optimization during training.
This strategy consists of convergence slope estimation, which estimates each type’s convergence rate through linear regression of loss, and adaptive reweighting, which integrates type balance score (TBS), type stability score (TSS), and adaptivity factor (AF), to dynamically balance type contributions and maintain stable optimization.

\vspace{-2mm}
{\flushleft\textbf{Convergence slope estimation}.}
We first estimate the convergence tendency of each type by analyzing the evolution of its loss.
Let $\{(k, y_k)\}_{k=t-N+1}^{t}$ denote the training-step indices and their corresponding normalized losses within a sliding window $[t-N+1,\,t]$, where $N$ is the window size and $t$ is the current iteration.
To capture short-term convergence dynamics, we apply a least-squares linear fitting over this window and define the regression coefficient $\alpha$ as the convergence slope, which measures the local rate of loss change.
The slope and intercept are calculated as:
\begin{eqnarray} 
\begin{aligned} 
\alpha &= \frac{\sum_{k=1}^{N}(k-\bar{k})(y_k-\bar{y})}{\sum_{k=1}^{N}(k-\bar{k})^2}, 
\qquad 
\beta = \bar{y} - \alpha\bar{k},
\end{aligned} 
\label{eq:linearfit} 
\end{eqnarray}
where $\bar{k}=\tfrac{1}{N}\sum_{k=t-N+1}^{t}k$ and $\bar{y}=\tfrac{1}{N}\sum_{k=t-N+1}^{t}y_k$ are the mean values within the window.
In this definition, $\alpha<0$ indicates a decreasing loss (converging), $\alpha>0$ implies divergence, and a larger $|\alpha|$ denotes a faster change rate.

\vspace{-2mm}
{\flushleft\textbf{Adaptive reweighting}.}
This estimated slope $\alpha$ serves as the foundation for computing the subsequent adaptive weighting indicators, namely type balance score (TBS), type stability score (TSS), and adaptivity factor (AF), which jointly determine how type weights evolve during training.
The goal of TBS is to assign smaller weights to types that converge faster and larger weights to those that converge more slowly, ensuring synchronized convergence among all types. For each type $i$ at iteration $t$, we define the TBS as
\begin{equation}
\mathrm{TBS}_i(t) = {softmax}_i\!\left( K \frac{\alpha_i(t)}{\sum_{i=1}^{K}|\alpha_i(t)|} \right), 
\label{eq:tbs} 
\end{equation}
where $\alpha_i(t)$ is the convergence slope of type $i$, ${softmax}_i$ operates across the type dimension, and $K$ denotes the total number of degradation types.
This normalization removes scale sensitivity and ensures that slower types obtain higher TBS values, thus receiving greater optimization emphasis.

However, relying solely on TBS is insufficient, since a diverging type with a large positive slope could still obtain an undesirably high score.
To mitigate this issue, we propose the TSS, which evaluates each type’s internal stability by considering its historical convergence behavior:
\begin{equation} 
\mathrm{TSS}_i(t) = {softmax}_i\!\left( - N \frac{\alpha_i(t)}{\sum_{k=t-N+1}^{t}|\alpha_i(k)|} \right), 
\label{eq:tss} 
\end{equation}
where normalization is performed along the temporal dimension $k$ from step $t\!-\!N\!+\!1$ to $t$, but subsequently applies the softmax function across the type dimension $i$. A higher TSS indicates stable and consistent convergence, whereas diverging types receive lower scores.

During early training, all types generally exhibit convergent behavior, in which case TBS should dominate the weighting process. 
As training proceeds, some types may begin to diverge and thus require greater influence from TSS. 
To achieve a transition between these two regimes, we define the AF that reflects the global divergence state:
\begin{equation}
\mathrm{AF}(t) = 
min\!\left(
t \cdot {softmax}_t
\!\left(
- \frac{\tau t \cdot \alpha_{\max}(t)}{\sum_{i=1}^{t}\alpha_{\max}(i)}
\right), 1
\right),
\label{eq:af}
\end{equation}
where $\alpha_{\max}(t)$ is the largest convergence slope across all types and ${softmax}_t$ is applied along the temporal dimension. 
The term $t$ prevents AF from monotonically decreasing as training progresses, while $\tau$ controls its sensitivity.
Finally, the type loss weights are defined as follows:
\begin{equation}
\omega_{i}(t) = 
\mathrm{AF}(t)\,\mathrm{TBS}(t) + 
\big(1-\mathrm{AF}(t)\big)\,\mathrm{TSS}(t), \quad i \in[1, K]
\label{eq:final_weight}
\end{equation}
where $\omega_{i}(t)$ is the dynamic weight vector across $K$ rain types.
This approach enables adaptive focus between inter-type balance and intra-type stability, ensuring steady convergence and improved generalization across all rain types.

\subsection{Asymmetric MoE architecture}
To address diverse rain degradation behaviors, we adopt an asymmetric MoE with the soft-MoE encoder and the hard-MoE decoder for adaptive and efficient processing.
{\flushleft\textbf{Soft-MoE encoder}.}
Given an input feature $x_{en}$, the soft router function $\mathcal{R}_{soft}$ generates continuous routing weights to adaptively combine multiple experts:
\begin{equation}
\mathcal{R}_{soft} = \sigma(\mathcal{W}(\varphi(x_{en}+\epsilon))),
\end{equation}
where $\mathcal{W}(\cdot)$ is a learnable projection, $\varphi(\cdot)$ is global average pooling, $\epsilon \sim \mathcal{N}(0,\sigma^2)$ is the Gaussian noise for load balancing, and $\sigma(\cdot)$ is the softmax.
The output $y_{en}$ combines all experts contributions through weighted aggregation:
\begin{equation}
y_{en} = \sum_{i=1}^{N} \mathcal{R}_{soft}^{i} \otimes y_{en}^{i},
\end{equation}
where $\otimes$ is the element-wise multiplication, $y_{en}^{i}$ is the feature produced by the $i$-th expert, $\mathcal{R}_{soft}^{i}$ is the routing weight of the $i$-th expert, and $N$ is the total number of experts.
{\flushleft\textbf{Hard-MoE decoder}.}
For the decoder input $x_{de}$, the hard router $\mathcal{R}_{hard}$ function selectively activates the most relevant experts via Top-$k$ routing. The process is defined as:
\begin{align} 
\mathcal{R}_{hard} &= \mathcal{T}_{k}(\sigma(\mathcal{W}(\varphi(x_{de}+\epsilon)))), \\ y_{de} &= \sum_{i=1}^{N} \mathcal{R}_{hard}^{i}  \cdot y_{de}^{i}, 
\end{align}
where $\mathcal{T}_{k}(\cdot)$ denotes the Top-$k$ selection operator and $y_{de}^{i}$ is the output of the $i$-th decoder expert.
This design enables the hard-MoE to efficiently allocate computation to the most relevant experts, focusing on structural and fine-grained detail reconstruction, and complementing the soft-MoE encoder for balanced expressiveness and efficiency.

\begin{table*}[t]
\caption{Quantitative evaluations on the proposed RainRAG dataset, where DRS, DRD, NRS, and NRD denote daytime rain streaks, daytime raindrops, nighttime rain streaks, and nighttime raindrops, respectively. The best and second-best values are \textbf{bold} and \underline{underlined}.}
\label{tab:RainRAG}
\vspace{-3mm}
\renewcommand\arraystretch{1.3}
    \resizebox{\linewidth}{!}{
        \begin{tabular}{lc|ccccccccccccccc}
            \toprule
            \multicolumn{2}{c|}{\multirow{2}{*}{Datasets}} & \multicolumn{15}{c}{\textbf{RainRAG Dataset}}                                                                                                                                                                                 \\ \cline{3-17} 
            \multicolumn{2}{c|}{}                          & \multicolumn{3}{c|}{\textbf{DRS}}                     & \multicolumn{3}{c|}{\textbf{DRD}}                     & \multicolumn{3}{c|}{\textbf{NRS}}                     & \multicolumn{3}{c|}{\textbf{NRD}}                     & \multicolumn{3}{c}{\textbf{Average}} \\ \hline
            \multicolumn{1}{l|}{Methods}    & Venue        & PSNR~$\uparrow$  & SSIM~$\uparrow$   & \multicolumn{1}{c|}{LPIPS~$\downarrow$}  & PSNR~$\uparrow$  & SSIM~$\uparrow$   & \multicolumn{1}{c|}{LPIPS~$\downarrow$}  & PSNR~$\uparrow$  & SSIM~$\uparrow$   & \multicolumn{1}{c|}{LPIPS~$\downarrow$}  & PSNR~$\uparrow$  & SSIM~$\uparrow$   & \multicolumn{1}{c|}{LPIPS~$\downarrow$}  & PSNR~$\uparrow$    & SSIM~$\uparrow$    & LPIPS~$\downarrow$   \\ \hline
            \multicolumn{1}{l|}{LQ}     & -    & 23.79 & 0.7514 & \multicolumn{1}{c|}{0.3207} & 20.79 & 0.7293 & \multicolumn{1}{c|}{0.2433} & 26.89 & 0.8480 & \multicolumn{1}{c|}{0.3583} & 21.05 & 0.5590 & \multicolumn{1}{c|}{0.6130} & 23.13   & 0.7220  & 0.3839  \\
            \multicolumn{1}{l|}{PReNet~\cite{ren2019progressive}}     & CVPR 2019    & 26.73 & 0.8233 & \multicolumn{1}{c|}{0.1885} & 21.45 & 0.7358 & \multicolumn{1}{c|}{0.2697} & 32.85 & 0.9520 & \multicolumn{1}{c|}{0.1957} & 24.29 & 0.8058 & \multicolumn{1}{c|}{0.3751} & 26.33   & 0.8292  & 0.2572  \\
            \multicolumn{1}{l|}{RCDNet~\cite{wang2020model}}     & CVPR 2020    & 25.27 & 0.7949 & \multicolumn{1}{c|}{0.2777} & 21.46 & 0.7295 & \multicolumn{1}{c|}{0.2778} & 28.44 & 0.9070 & \multicolumn{1}{c|}{0.2996} & 20.62 & 0.5832 & \multicolumn{1}{c|}{0.5931} & 23.95   & 0.7537  & 0.3620  \\
            \multicolumn{1}{l|}{MPRNet~\cite{zamir2021multi}}     & CVPR 2021    & 27.99 & 0.8242 & \multicolumn{1}{c|}{\underline{0.1701}} & 22.50 & 0.7264 & \multicolumn{1}{c|}{0.3308} & 33.75 & 0.9541 & \multicolumn{1}{c|}{0.1834} & 24.46 & 0.8118 & \multicolumn{1}{c|}{0.3634} & 27.17   & 0.8291  & 0.2619  \\
            \multicolumn{1}{l|}{Restormer~\cite{zamir2022restormer}}  & CVPR 2022    & 28.45 & 0.8298 & \multicolumn{1}{c|}{0.1725} & \underline{23.36} & 0.7547 & \multicolumn{1}{c|}{0.2323} & 33.92 & 0.9541 & \multicolumn{1}{c|}{0.1830} & \underline{25.85} & 0.8234 & \multicolumn{1}{c|}{0.3452} & 27.89   & 0.8405  & \underline{0.2332}  \\
            \multicolumn{1}{l|}{IDT~\cite{xiao2022image}}        & TPAMI 2022   & 26.97 & 0.8279 & \multicolumn{1}{c|}{0.1960} & 23.17 & \underline{0.7648} & \multicolumn{1}{c|}{\textbf{0.2300}} & 33.05 & 0.9470 & \multicolumn{1}{c|}{0.2132} & 25.20 & 0.7963 & \multicolumn{1}{c|}{0.3718} & 27.10   & 0.8340  & 0.2527  \\
            \multicolumn{1}{l|}{DRSformer~\cite{chen2023learning}}  & CVPR 2023    & 28.08 & 0.8253 & \multicolumn{1}{c|}{0.1960} & 22.93 & 0.7499 & \multicolumn{1}{c|}{0.2407} & 32.86 & 0.9457 & \multicolumn{1}{c|}{0.2202} & 24.99 & 0.7953 & \multicolumn{1}{c|}{0.3702} & 27.22   & 0.8291  & 0.2568  \\
            \multicolumn{1}{l|}{RLP~\cite{zhang2023learning}}        & ICCV 2023    & 28.01 & 0.8290 & \multicolumn{1}{c|}{0.2339} & 22.42 & 0.7353 & \multicolumn{1}{c|}{0.3029} & 32.06 & 0.9438 & \multicolumn{1}{c|}{0.1971} & 23.48 & 0.7638 & \multicolumn{1}{c|}{0.4282} & 26.49   & 0.8180  & 0.2905  \\
            \multicolumn{1}{l|}{MSDT~\cite{chen2024rethinking}}       & AAAI 2024    & \underline{28.60} & \textbf{0.8414} & \multicolumn{1}{c|}{\textbf{0.1576}} & 23.31 & 0.7487 & \multicolumn{1}{c|}{0.2587} & \underline{34.56} & \underline{0.9591} & \multicolumn{1}{c|}{\underline{0.1677}} & 25.28 & 0.8148 & \multicolumn{1}{c|}{0.3555} & \underline{27.94}   & 0.8410  & 0.2349  \\
            \multicolumn{1}{l|}{NeRD-Rain~\cite{chen2024bidirectional}}  & CVPR 2024    & 28.11 & 0.8309 & \multicolumn{1}{c|}{0.1720} & 23.30 & 0.7513 & \multicolumn{1}{c|}{\underline{0.2318}} & 33.88 & 0.9511 & \multicolumn{1}{c|}{0.1964} & 25.31 & 0.8027 & \multicolumn{1}{c|}{0.3571} & 27.65   & 0.8340  & 0.2393  \\
            \multicolumn{1}{l|}{URIR~\cite{yan2025towards}}       & AAAI 2025    & 28.29 & 0.8386 & \multicolumn{1}{c|}{0.2001} & 23.19 & 0.7483 & \multicolumn{1}{c|}{0.2815} & 34.32 & 0.9570 & \multicolumn{1}{c|}{0.1860} & 25.82 & \underline{0.8263} & \multicolumn{1}{c|}{\underline{0.3390}} & 27.91   & \underline{0.8425}  & 0.2516  \\
            \multicolumn{1}{l|}{ UniRain (Ours)}       & -            & \textbf{29.58} & \underline{0.8400} & \multicolumn{1}{c|}{0.1814} & \textbf{24.71} & \textbf{0.7687} & \multicolumn{1}{c|}{0.2408} & \textbf{35.23}   & \textbf{0.9592}  & \multicolumn{1}{c|}{\textbf{0.1617}} & \textbf{26.21} & \textbf{0.8380} & \multicolumn{1}{c|}{\textbf{0.3275}}  &  \textbf{28.93} & \textbf{0.8515} & \textbf{0.2279}  \\  \bottomrule
        \end{tabular}
    }
    \vspace{-3mm}
\end{table*}

\begin{figure*}[t]
	\footnotesize
	\begin{center}
		\begin{tabular}{c c c c c c c c}
			\multicolumn{3}{c}{\multirow{5}*[48pt]{
            \hspace{-2.5mm} \includegraphics[width=0.42\linewidth,height=0.245\linewidth]{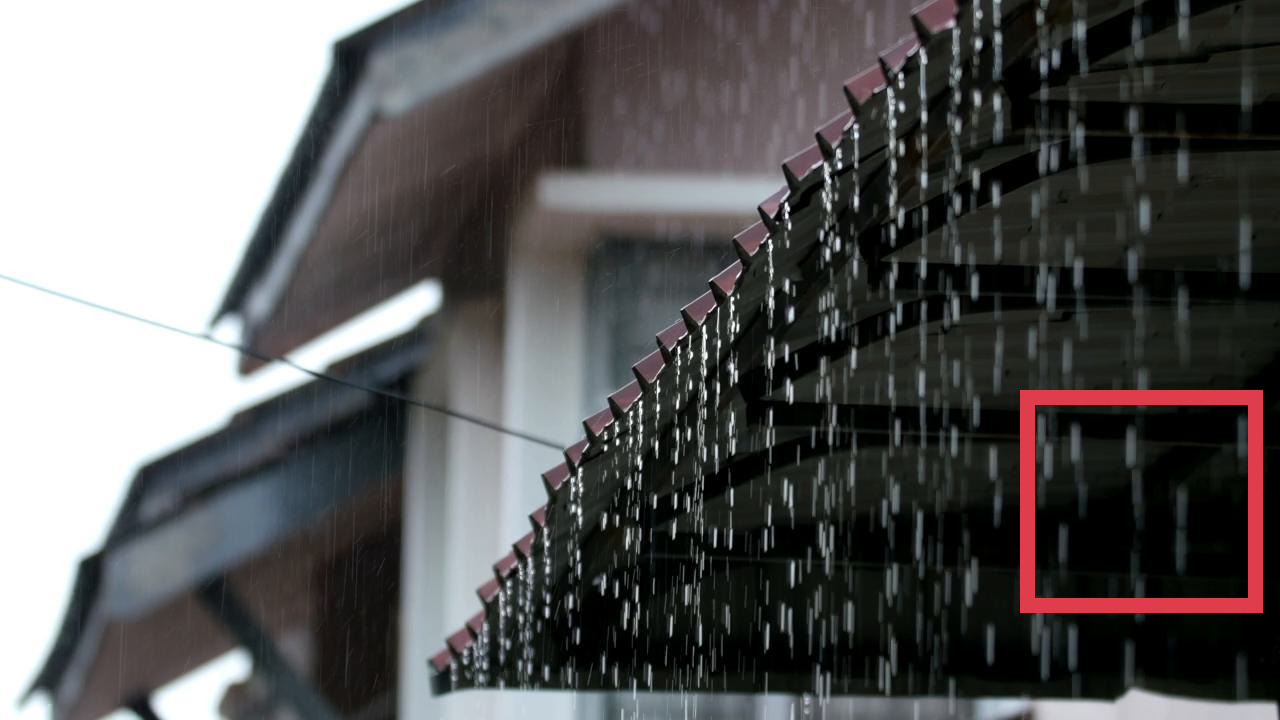}}}
            & \hspace{-4.0mm} \includegraphics[width=0.11\linewidth]{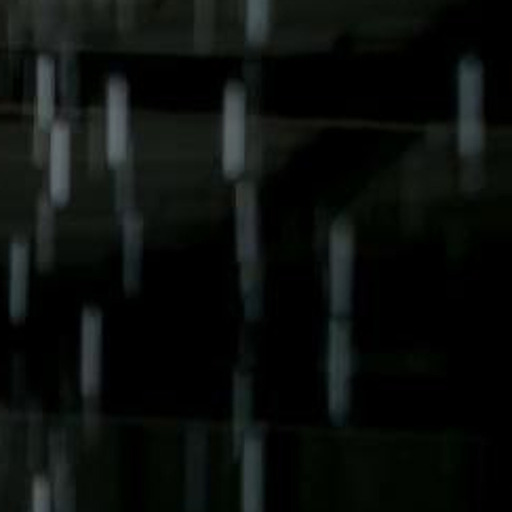}
            & \hspace{-4.0mm} \includegraphics[width=0.11\linewidth]{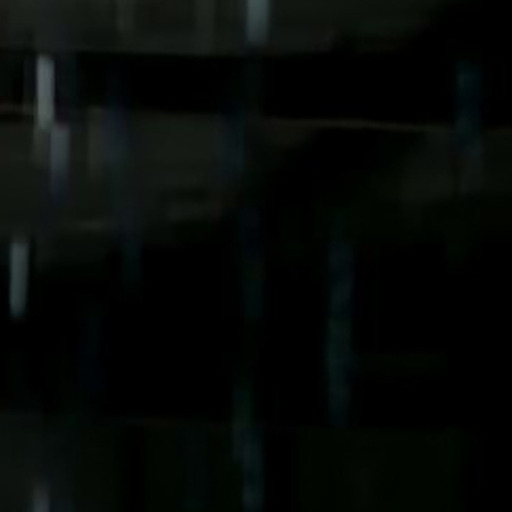} 
            & \hspace{-4.0mm} \includegraphics[width=0.11\linewidth]{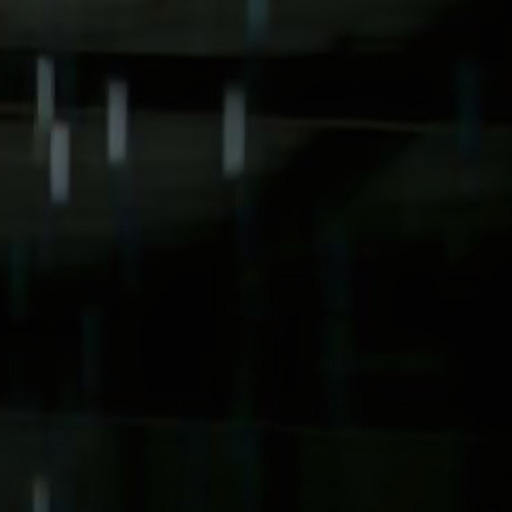} 
            & \hspace{-4.0mm} \includegraphics[width=0.11\linewidth]{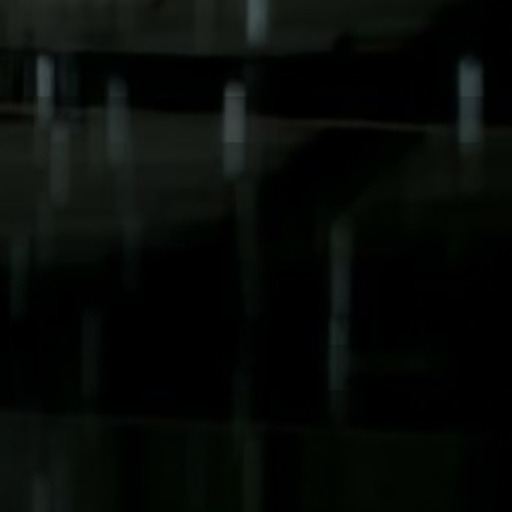} 
            & \hspace{-4.0mm} \includegraphics[width=0.11\linewidth]{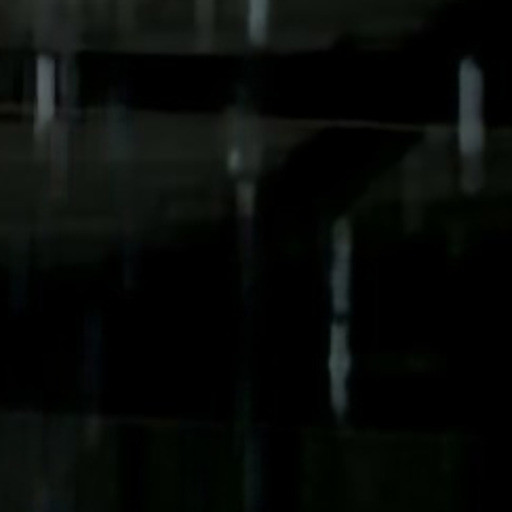} 
              \\
		\multicolumn{3}{c}{~}     
        & \hspace{-4.0mm} LQ patch 
        & \hspace{-4.0mm} PReNet~\cite{ren2019progressive}
        & \hspace{-4.0mm} RCDNet~\cite{wang2020model}
        & \hspace{-4.0mm} IDT~\cite{xiao2022image} 
        & \hspace{-4.0mm} RLP~\cite{zhang2023learning} \\  	
	\multicolumn{3}{c}{~} 
        & \hspace{-4.0mm} \includegraphics[width=0.11\linewidth]{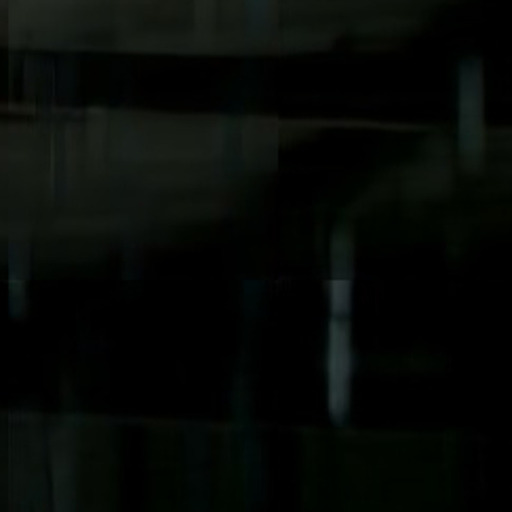}
        & \hspace{-4.0mm} \includegraphics[width=0.11\linewidth]{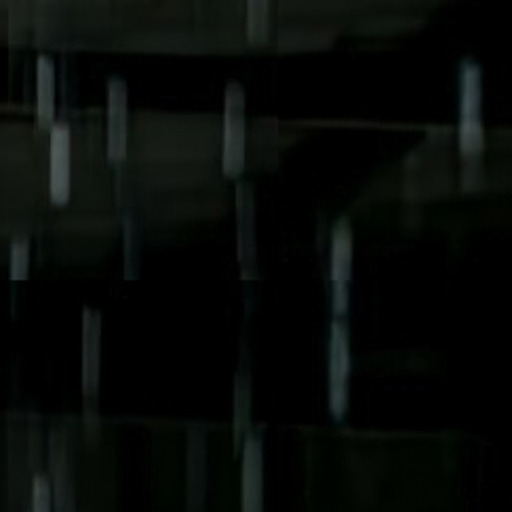} 
        & \hspace{-4.0mm} \includegraphics[width=0.11\linewidth]{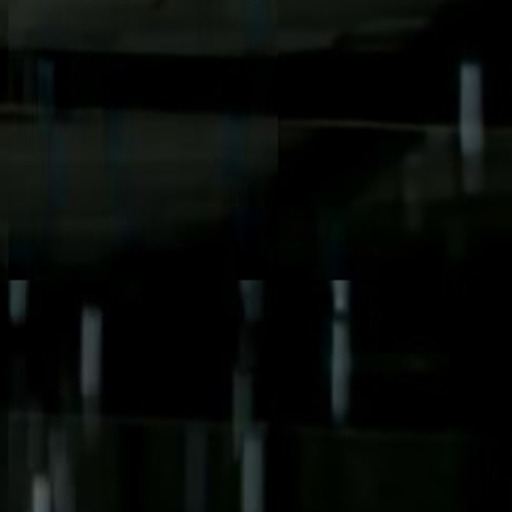} 
        & \hspace{-4.0mm} \includegraphics[width=0.11\linewidth]{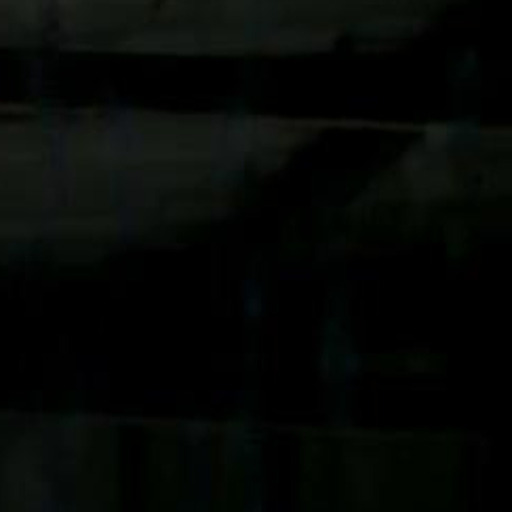} 
        & \hspace{-4.0mm} \includegraphics[width=0.11\linewidth]{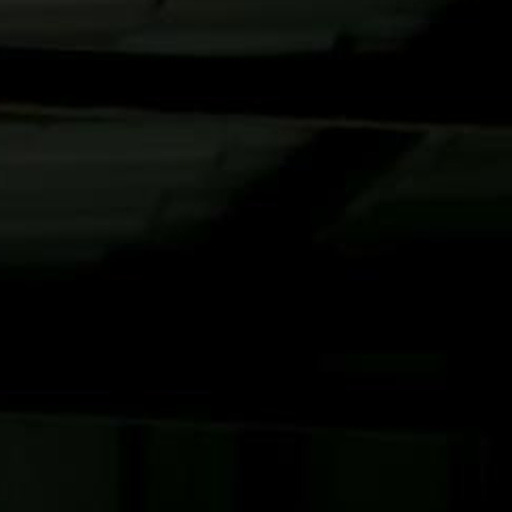} 

        \\
	\multicolumn{3}{c}{\hspace{-4.0mm} Rainy image from RainRAG-NRS} 
        & \hspace{-4.0mm} MSDT~\cite{chen2024rethinking}
        & \hspace{-4.0mm} NeRD-Rain~\cite{chen2024bidirectional}
        & \hspace{-4.0mm} URIR~\cite{yan2025towards}
        & \hspace{-4.0mm} \textbf{UniRain}
        & \hspace{-4.0mm} GT patch \\ 	

        \multicolumn{3}{c}{\multirow{5}*[48pt]{
            \hspace{-2.5mm} \includegraphics[width=0.42\linewidth,height=0.245\linewidth]{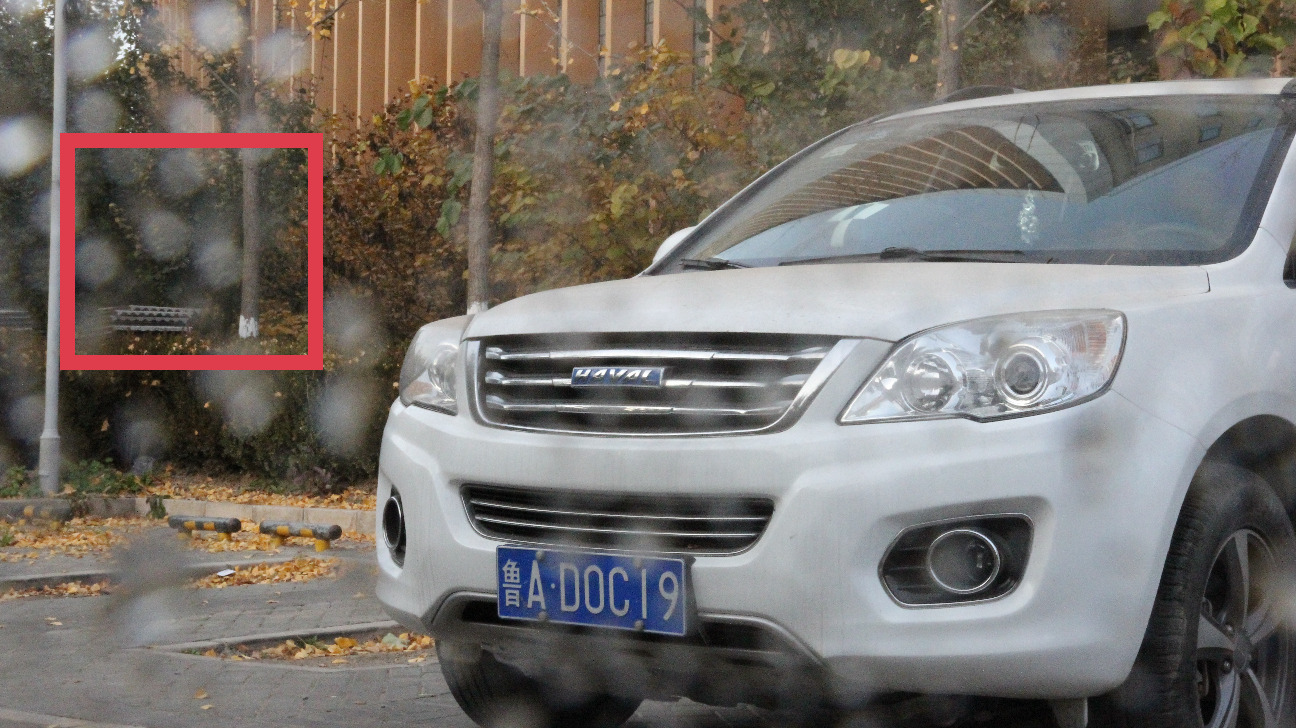}}}
            & \hspace{-4.0mm} \includegraphics[width=0.11\linewidth]{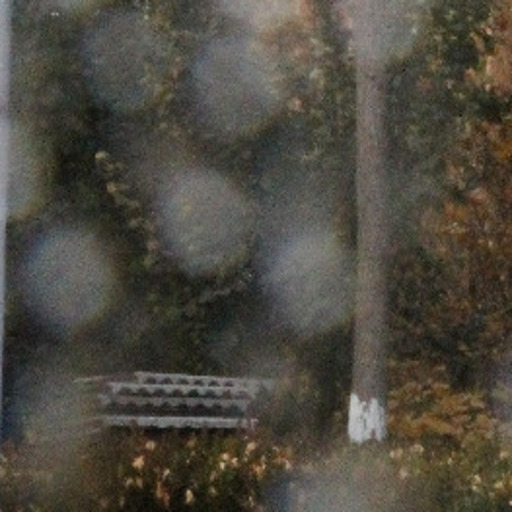}
            & \hspace{-4.0mm} \includegraphics[width=0.11\linewidth]{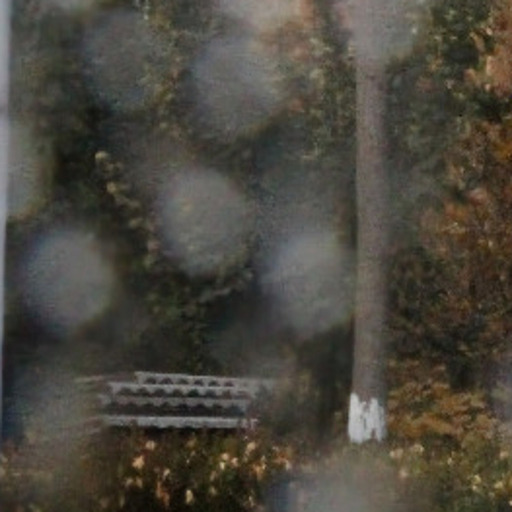} 
            & \hspace{-4.0mm} \includegraphics[width=0.11\linewidth]{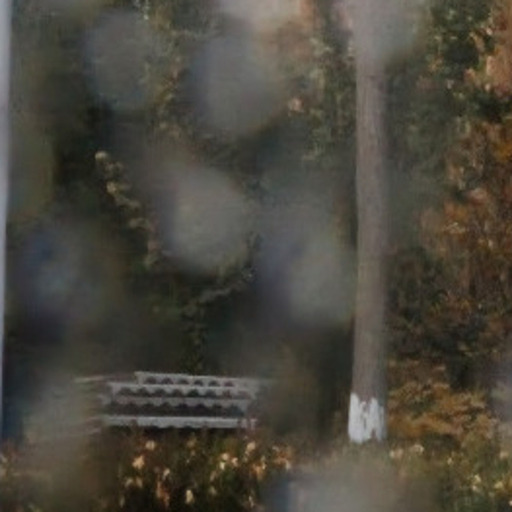} 
            & \hspace{-4.0mm} \includegraphics[width=0.11\linewidth]{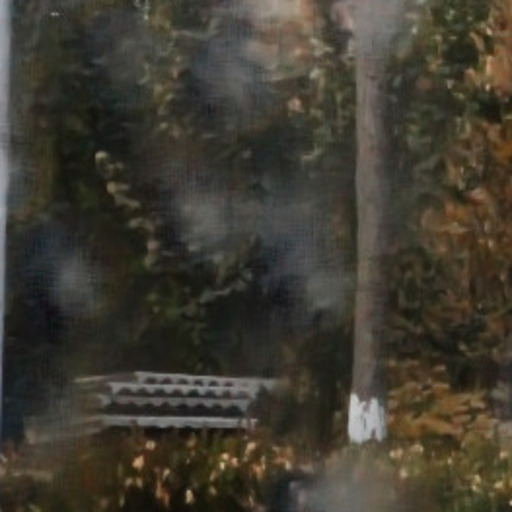} 
            & \hspace{-4.0mm} \includegraphics[width=0.11\linewidth]{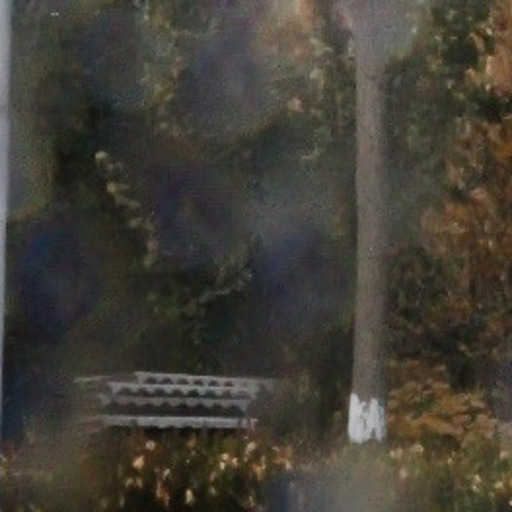} 
              \\
		\multicolumn{3}{c}{~}                                   & \hspace{-4.0mm} LQ patch
        & \hspace{-4.0mm} PReNet~\cite{ren2019progressive}
        & \hspace{-4.0mm} MPRNet~\cite{zamir2021multi}
        & \hspace{-4.0mm} Restormer~\cite{zamir2022restormer}	
        & \hspace{-4.0mm} DRSformer~\cite{chen2023learning} \\
	\multicolumn{3}{c}{~} 
        & \hspace{-4.0mm} \includegraphics[width=0.11\linewidth]{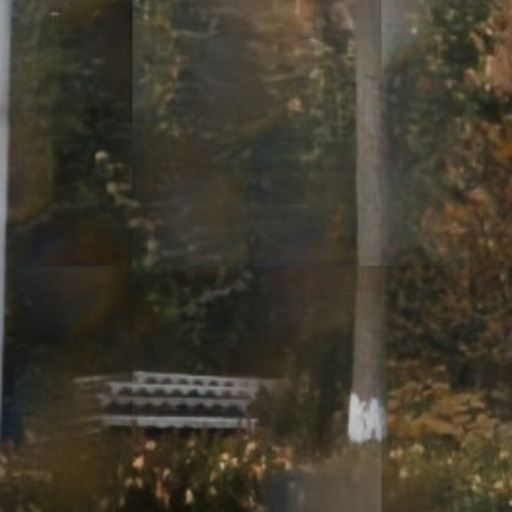} 
        & \hspace{-4.0mm} \includegraphics[width=0.11\linewidth]{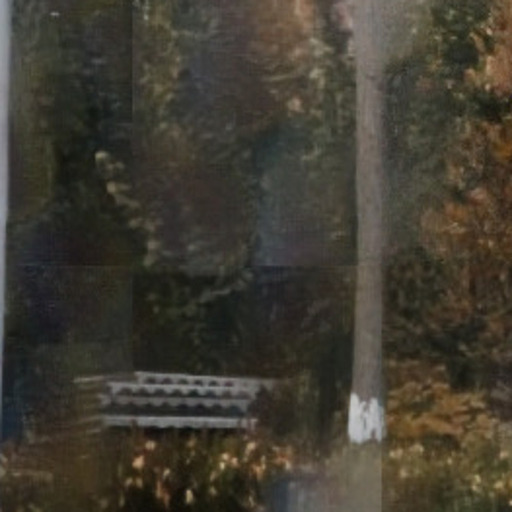} 
        & \hspace{-4.0mm} \includegraphics[width=0.11\linewidth]{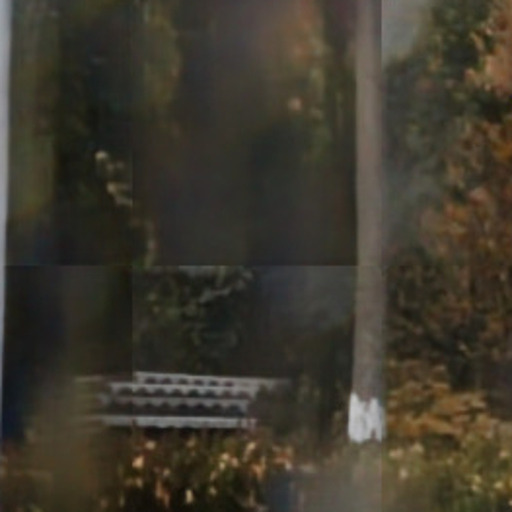} 
        & \hspace{-4.0mm} \includegraphics[width=0.11\linewidth]{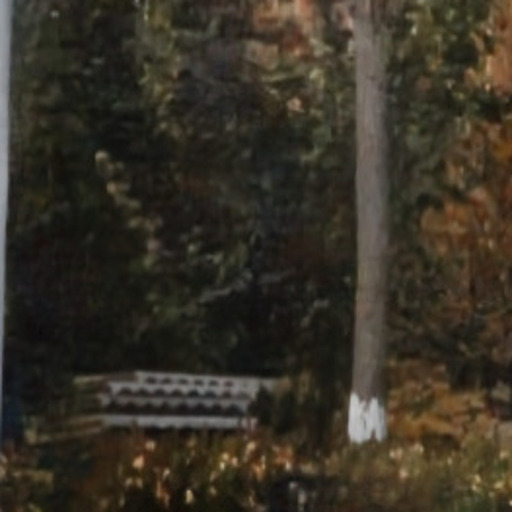} 
        & \hspace{-4.0mm} \includegraphics[width=0.11\linewidth]{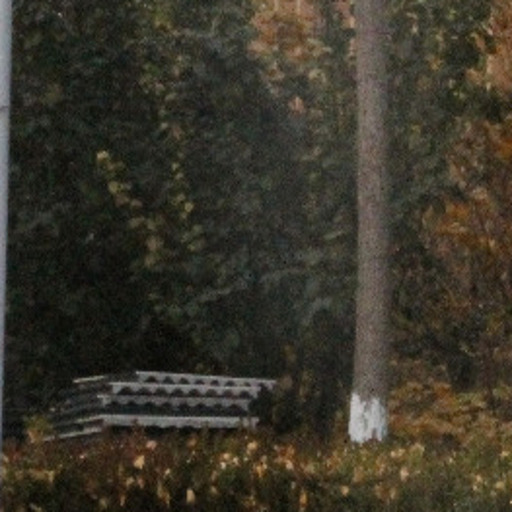}
        \\
	\multicolumn{3}{c}{\hspace{-4.0mm} Rainy image from RainDS-real-RD~\cite{quan2021removing}} 
        & \hspace{-4.0mm} MSDT~\cite{chen2024rethinking}
        & \hspace{-4.0mm} NeRD-Rain~\cite{chen2024bidirectional}
        & \hspace{-4.0mm} URIR~\cite{yan2025towards}
        & \hspace{-4.0mm} \textbf{UniRain}
        & \hspace{-4.0mm} GT patch
        \\ 
		\end{tabular}
	\end{center}
	\vspace{-6mm}
	\caption{Visual comparison of image deraining results on the RainRAG-NRS and RainDS-real-RD datasets.~Zoom in for a better view.}
	\label{fig:visual}
	\vspace{-5mm}
\end{figure*}


\begin{table*}[htb]
\caption{Quantitative evaluations on the real-world public benchmarks.~The best and second-best values are \textbf{bold} and \underline{underlined}.}
\label{tab:public_dataset}
\vspace{-3mm}
\renewcommand\arraystretch{1.3}
    \resizebox{\linewidth}{!}{
        \begin{tabular}{l|c|ccccccccccc}
            \toprule
            Datasets                         & Metrics & PReNet~\cite{ren2019progressive} & RCDNet~\cite{wang2020model} & MPRNet~\cite{zamir2021multi} & IDT~\cite{xiao2022image}    & Restormer~\cite{zamir2022restormer} & DRSformer~\cite{chen2023learning} & RLP~\cite{zhang2023learning}    & MSDT~\cite{chen2024rethinking}   & NeRD-Rain~\cite{chen2024bidirectional} & URIR~\cite{yan2025towards}   & UniRain (Ours)    \\ \hline
            \multirow{2}{*}{RealRain-1k-L~\cite{li2022toward}}   & PSNR~$\uparrow$    & 28.80  & 28.58  & 33.89  & 33.71  & 32.44     & 33.65     & 32.02  & 33.56  & \underline{33.98}     & 33.54  & \textbf{36.51}   \\
                                             & SSIM~$\uparrow$    & 0.9307 & 0.9202 & 0.9701 & 0.9664 & 0.9644    & 0.9625    & 0.9558 & 0.9640 & \underline{0.9710}    & 0.9686 & \textbf{0.9778}  \\ \hline
            \multirow{2}{*}{RealRain-1k-H~\cite{li2022toward}}   & PSNR~$\uparrow$    & 25.49  & 25.46  & 30.83  & 30.37  & 28.80     & 30.35     & 29.01  & 30.91  & \underline{31.37}     & 31.10  & \textbf{33.74}   \\
                                             & SSIM~$\uparrow$    & 0.8820 & 0.8613 & 0.9542 & 0.9441 & 0.9389    & 0.9360    & 0.9309 & 0.9451 & \underline{0.9583}    & 0.9550 & \textbf{0.9672}  \\ \hline
            \multirow{2}{*}{RainDS-real-RD~\cite{quan2021removing}}  & PSNR~$\uparrow$    & 19.08  & 19.37  & 19.48  & 20.35  & 20.74     & 20.15     & 20.86  & 20.72  & \underline{21.01}     & 20.44  & \textbf{22.07}   \\
                                             & SSIM~$\uparrow$    & 0.6020 & 0.5937 & 0.5957 & 0.6237 & \underline{0.6346}    & 0.6133    & 0.6136 & 0.6205 & 0.6155    & 0.5945 & \textbf{0.6436}  \\ \hline
            \multirow{2}{*}{RainDS-real-RDS~\cite{quan2021removing}} & PSNR~$\uparrow$    & 18.21  & 18.57  & 18.74  & 18.98  & 19.13     & 18.79     & 19.85  & 19.36  & \underline{19.93}     & 19.87  & \textbf{20.61}   \\
                                             & SSIM~$\uparrow$    & 0.5665 & 0.5572 & 0.5681 & 0.5833 & \underline{0.5876}    & 0.5712    & 0.5762 & 0.5830 & 0.5839    & 0.5689 & \textbf{0.5881}  \\ \hline
            \multirow{2}{*}{WeatherBench-rain~\cite{guan2025weatherbench}}    & PSNR~$\uparrow$    & 31.18  & 27.91  & 32.35  & 32.05  & 32.91     & 31.90     & 31.48  & \underline{33.56}  & 32.77     & 33.50  & \textbf{34.25}   \\
                                             & SSIM~$\uparrow$    & 0.9364 & 0.8880 & 0.9352 & 0.9355 & 0.9372    & 0.9281    & 0.9330 & \textbf{0.9453} & 0.9372    & 0.9435 & \underline{0.9448}  \\ \hline
            \multirow{2}{*}{Average}         & PSNR~$\uparrow$    & 24.55  & 23.98  & 27.06  & 27.09  & 26.80     & 26.97     & 26.64  & 27.62  & \underline{27.81}     & 27.69  & \textbf{29.42}   \\
                                             & SSIM~$\uparrow$    & 0.7835 & 0.7641 & 0.8047 & 0.8106 & 0.8125    & 0.8022    & 0.8019 & 0.8116 & \underline{0.8132}    & 0.8061 & \textbf{0.8222} \\ \bottomrule
        \end{tabular}
    }
    \vspace{-2mm}
\end{table*}

\section{Experiments}
\label{sec:experiments}

\subsection{Experimental settings}
{\flushleft\textbf{Training dataset}.}~Through the RAG-based dataset distillation pipeline, we construct RainRAG, a high-quality dataset distilled from large-scale (\textit{i.e.}, $>$ 2,000k pairs) public synthetic and real-world deraining datasets~\cite{yan2025towards, yang2017deep, jiang2020multi, qian2018attentive, quan2021removing, zhang2022gtav, guan2025cstnet, wang2019spatial, ba2022not, guo2023sky, wen2023video, chang2024uav}.
Our pipeline retains approximately 2.6$\%$ of the original data, ensuring only high-quality samples for model training.
The training set consists of 52,869 pairs and divided into four representative subsets: daytime rain streaks (DRS), daytime raindrops (DRD), nighttime rain streaks (NRS), and nighttime raindrops (NRD).

\vspace{-2mm}
{\flushleft\textbf{Test datasets}.}~To comprehensively evaluate our model’s capability in unified image deraining, we obtain 400 test image pairs using the same filtering process and evenly divide them into four subsets: DRS, DRD, NRS, and NRD. In addition, we evaluate our method alongside other comparison methods on several real-world benchmarks (\textit{i.e.}, RealRain-1k~\cite{li2022toward}, RainDS-real~\cite{quan2021removing}, and WeatherBench~\cite{guan2025weatherbench}).

\vspace{-2mm}
{\flushleft\textbf{Implementation details}.}
We train our proposed UniRain on 4 NVIDIA GeForce RTX 4090 GPUs using the AdamW optimizer with default parameters.
During training, images are randomly cropped into $128\times128$ patches with a batch size of 8.
The initial learning rate is set to $2\times10^{-4}$ and gradually reduces to $1\times10^{-6}$ using a cosine annealing scheduler. 
For the asymmetric MoE architecture, each expert is designed with a distinct self-attention capacity, enabling diverse feature modeling across varying attention complexities.
We train the network for a total of 300,000 iterations.
For multi-objective reweighted optimization, we set the step window size to 10 and the sensitivity parameter $\tau$ to 5. 

\subsection{Comparison with the state-of-the-art}
We compare our proposed UniRain with 10 image deraining technologies, including PReNet~\cite{ren2019progressive}, RCDNet~\cite{wang2020model}, MPRNet~\cite{zamir2021multi}, Restormer~\cite{zamir2022restormer}, IDT~\cite{xiao2022image}, DRSformer~\cite{chen2023learning}, RLP~\cite{zhang2023learning}, MSDT~\cite{chen2024rethinking}, NeRD-Rain~\cite{chen2024bidirectional}, and URIR~\cite{yan2025towards}.

\vspace{-2mm}
{\flushleft\textbf{Evaluation on the RainRAG dataset}.}~Table~\ref{tab:RainRAG} shows that the proposed UniRain achieves the best average performance across four rain degradation types. Compared with Restormer~\cite{zamir2022restormer} and NeRD-Rain~\cite{chen2024bidirectional}, it achieves improvements of 1.35 dB and 1.41 dB in PSNR on the DRD subset, respectively. As shown in Figure~\ref{fig:visual}, we further present the visual comparisons under various rain degradations, where our method delivers superior deraining performance. In contrast, other methods still exhibit rain streak residuals and fail to recover background details affected by raindrops.

\begin{figure*}[htbp]
	\footnotesize
	\centering
    	\begin{tabular}{ccccccc}
            \hspace{-2.5mm}
    		\includegraphics[width=0.138\textwidth]{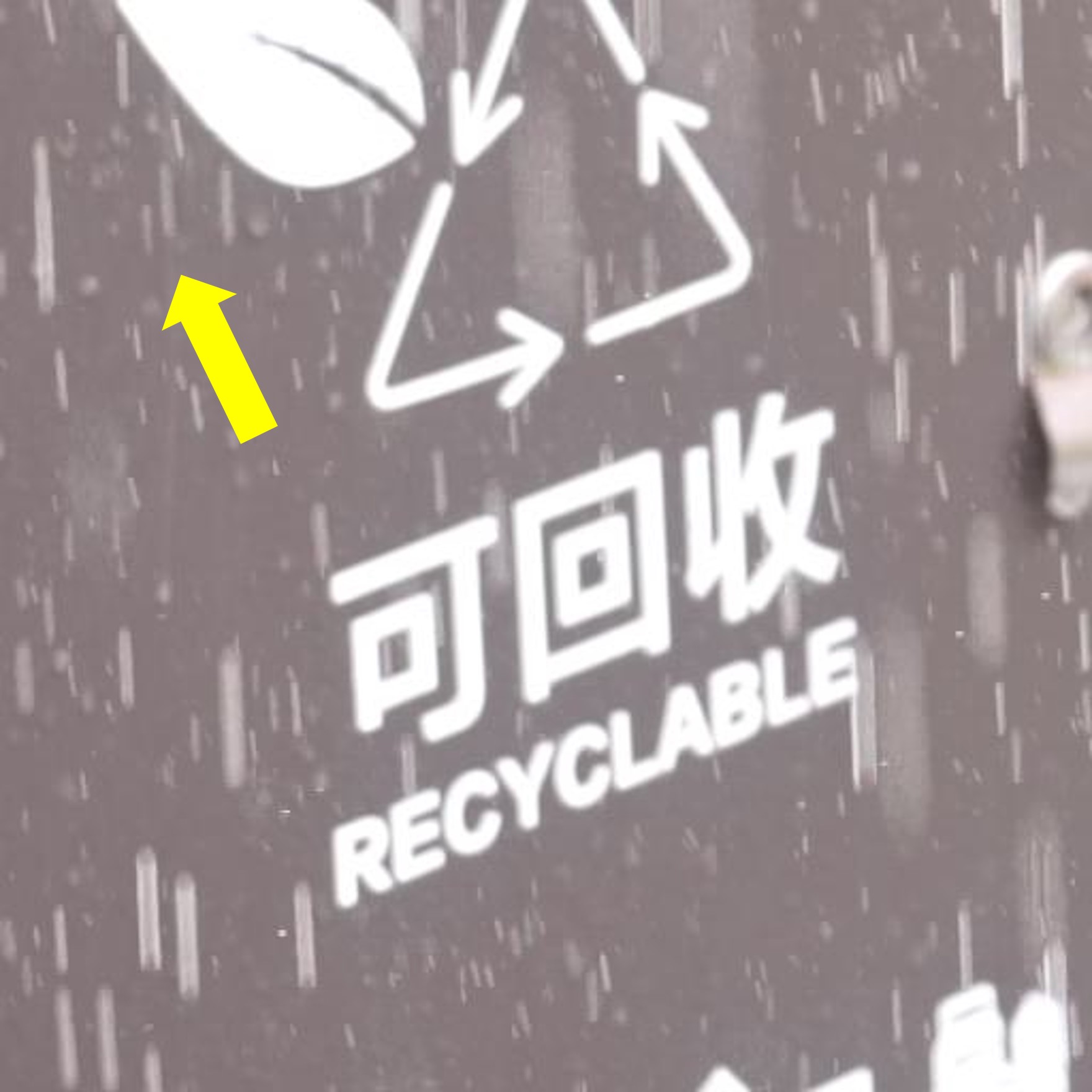}&\hspace{-4.1mm}
    		\includegraphics[width=0.138\textwidth]{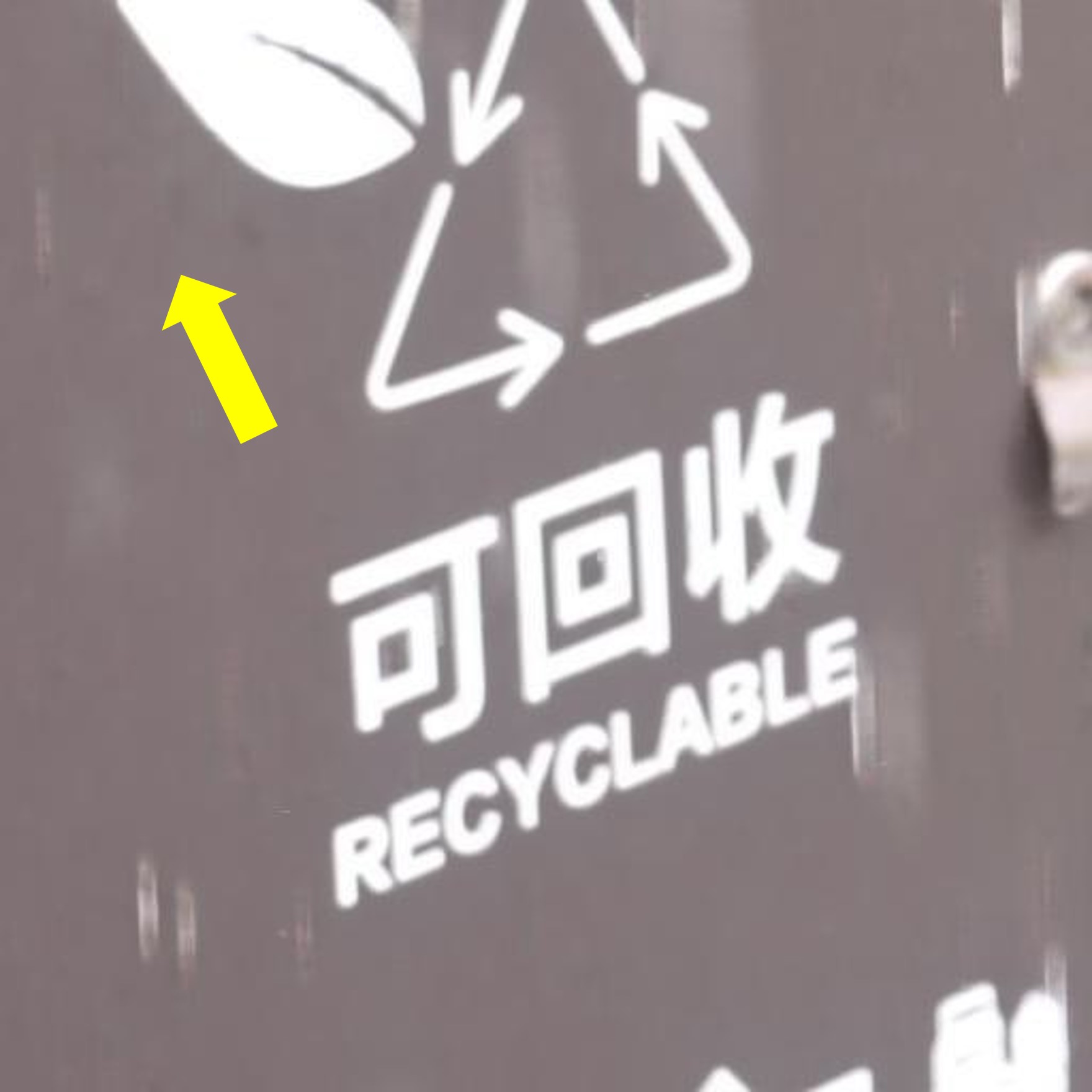}&\hspace{-4.1mm}
    		\includegraphics[width=0.138\textwidth]{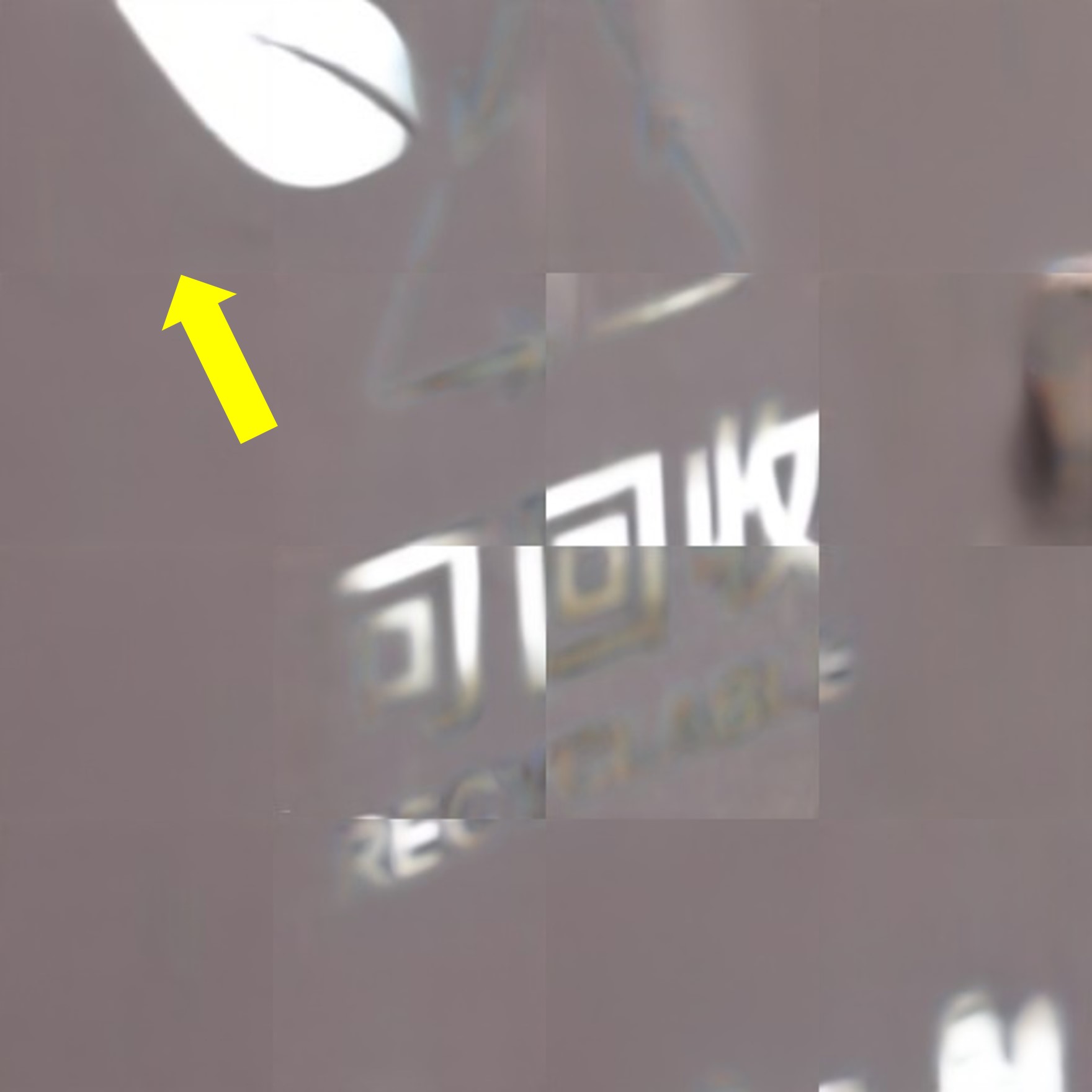}&\hspace{-4.1mm}
    		\includegraphics[width=0.138\textwidth]{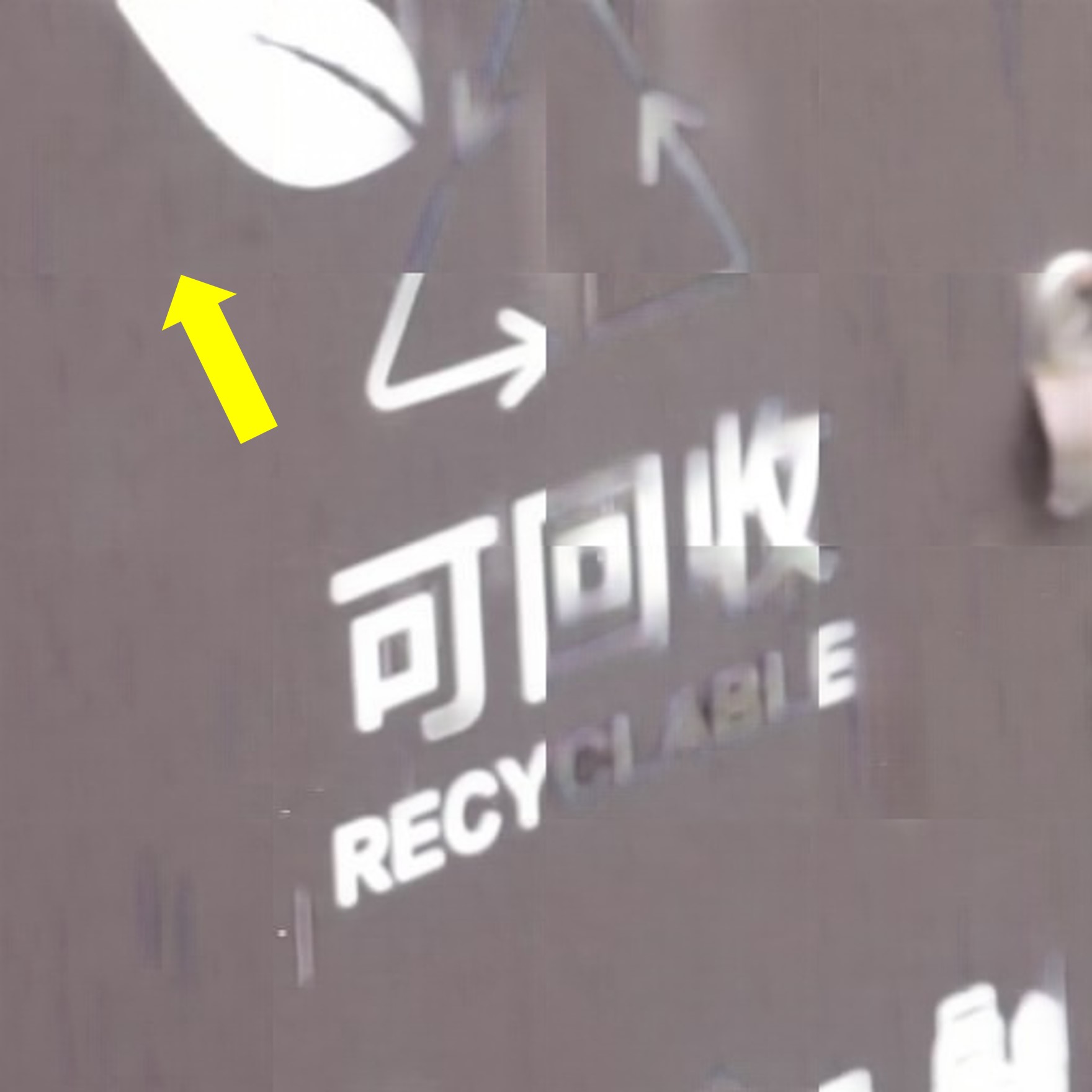}&\hspace{-4.1mm}
    		\includegraphics[width=0.138\textwidth]{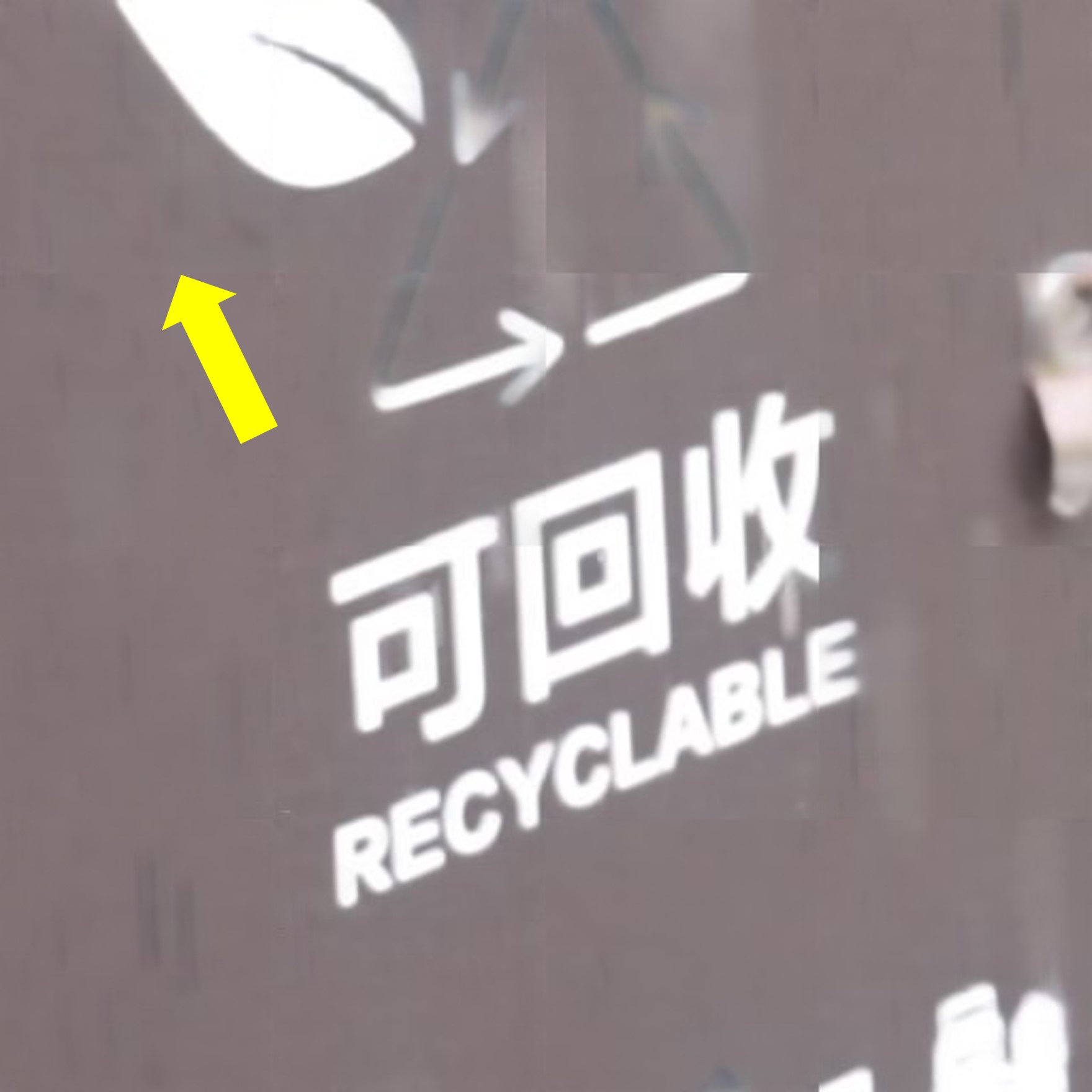}&\hspace{-4.1mm}
    		\includegraphics[width=0.138\textwidth]{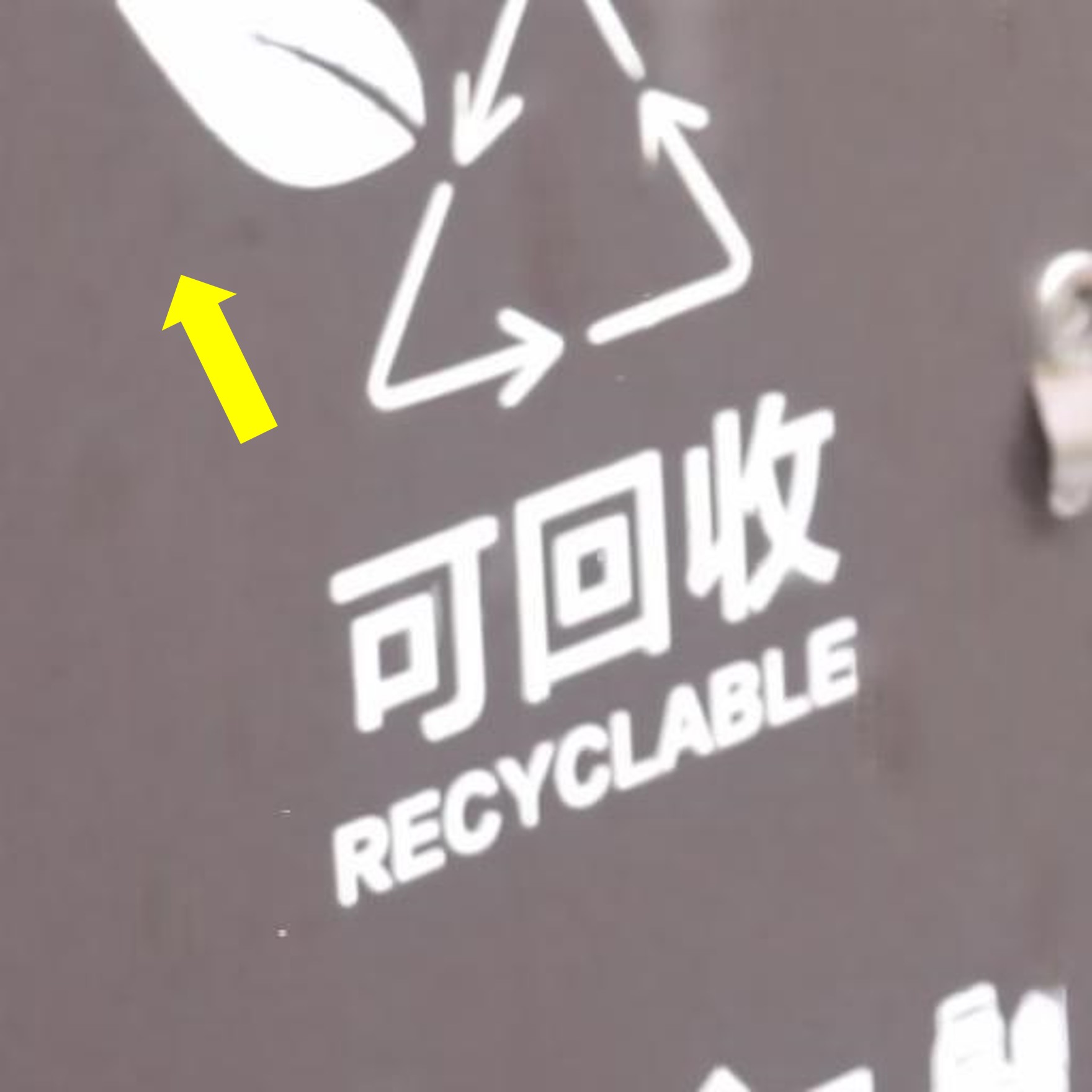}&\hspace{-4.1mm}
    		\includegraphics[width=0.138\textwidth]{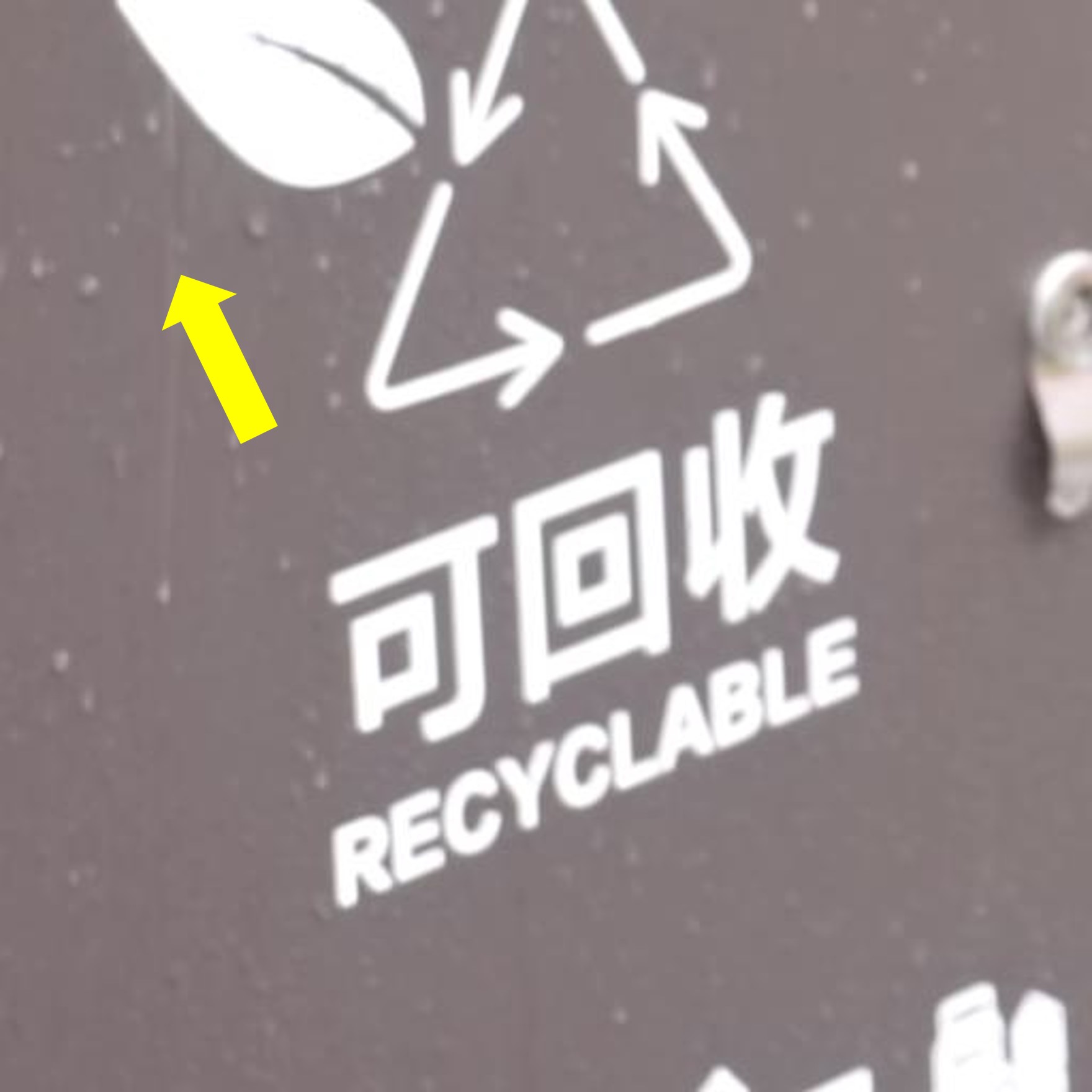} \\
    		LQ &\hspace{-4.1mm}  Restormer~\cite{zamir2022restormer}   &\hspace{-4.1mm}  MSDT~\cite{chen2024rethinking} &\hspace{-4.1mm} NeRD-Rain~\cite{chen2024bidirectional} &\hspace{-4.1mm}  URIR~\cite{yan2025towards} &\hspace{-4.1mm}  \textbf{UniRain} &\hspace{-4.1mm} GT\\
    	\end{tabular}
	\vspace{-3mm}
	\caption{Visual comparison of image restoration results on the real-world benchmark (\textit{i.e.}, WeatherBench~\cite{guan2025weatherbench}).~Compared to the state-of-the-art methods, our UniRain restores a high-quality image with clear details, even outperforming the GT by removing residual raindrops.}
	\label{fig:real-world}
	\vspace{-4mm}
\end{figure*}

\vspace{-2mm}
{\flushleft\textbf{Evaluation on real-world public benchmarks}.}~For further general verification in practical applications, we conduct comparisons with other methods on the real-world rainy datasets. As shown in Table~\ref{tab:public_dataset}, our method achieves the best overall performance, outperforming URIR~\cite{yan2025towards} by 1.73 dB in PSNR on average. Figure~\ref{fig:real-world} shows that our method not only removes real and complex rain streaks but also restores clear details, even achieving visually better results than the GT by eliminating residual raindrops contained in it.

\begin{figure}[!t]
    \centering
    \begin{subfigure}{0.32\linewidth}
        \centering
        \includegraphics[width=\linewidth]{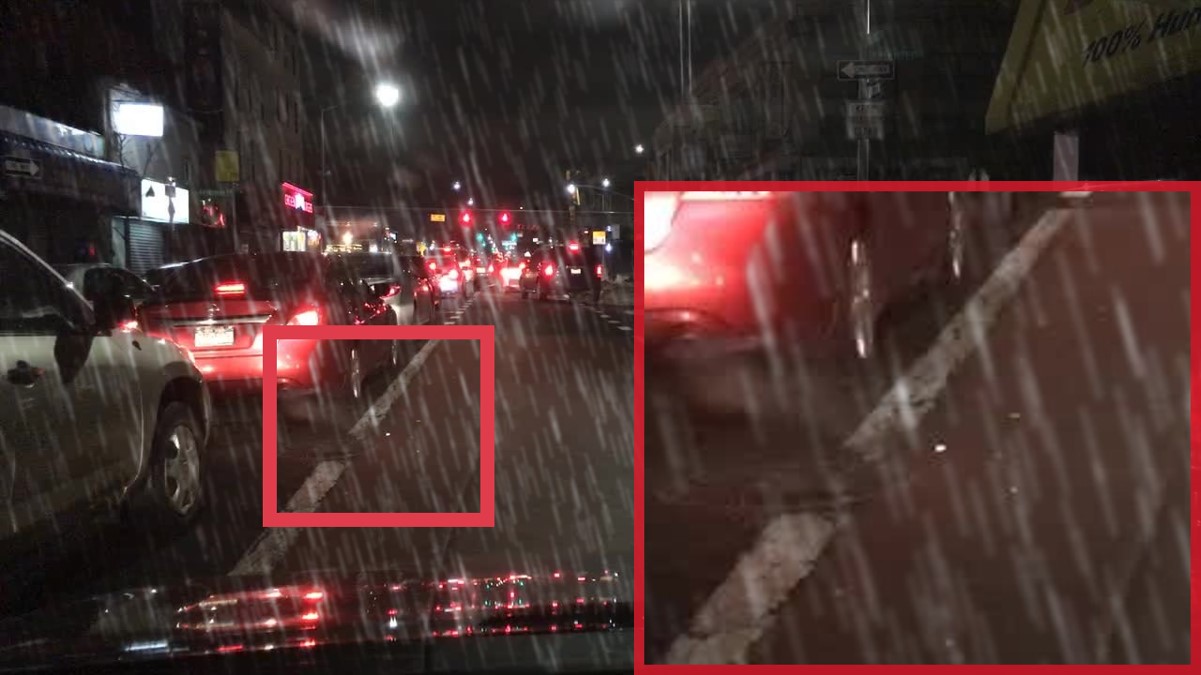}
    \end{subfigure}
    \begin{subfigure}{0.32\linewidth}
        \centering
        \includegraphics[width=\linewidth]{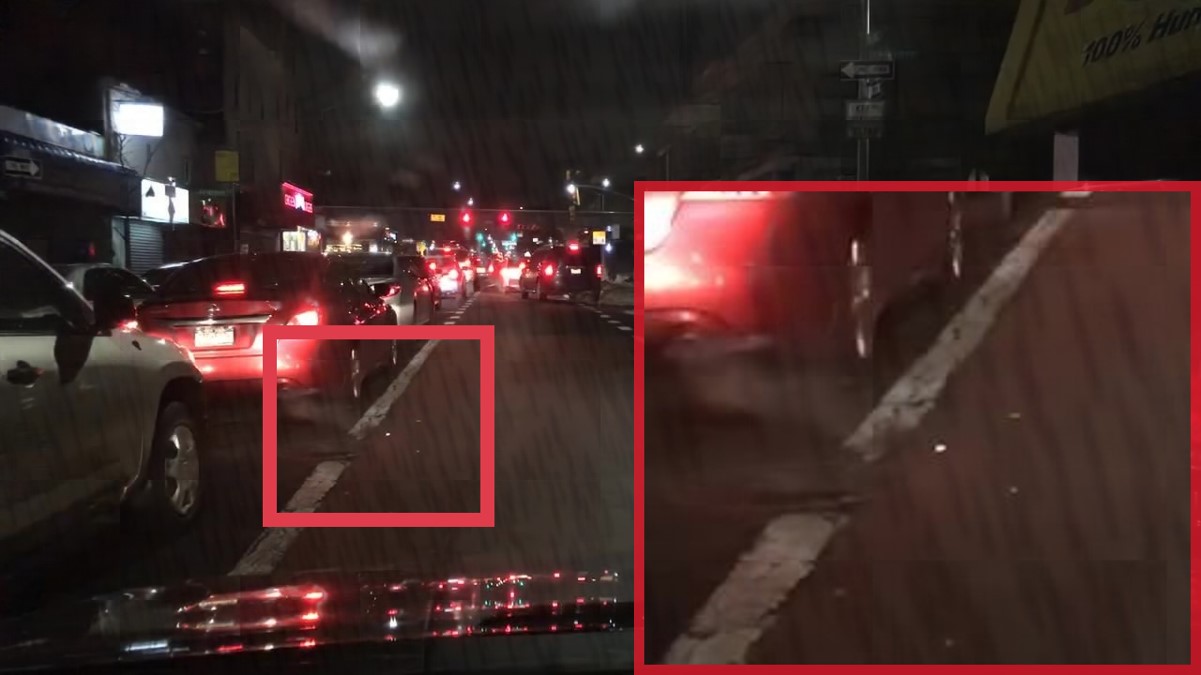}
    \end{subfigure}
    \begin{subfigure}{0.32\linewidth}
        \centering
        \includegraphics[width=\linewidth]{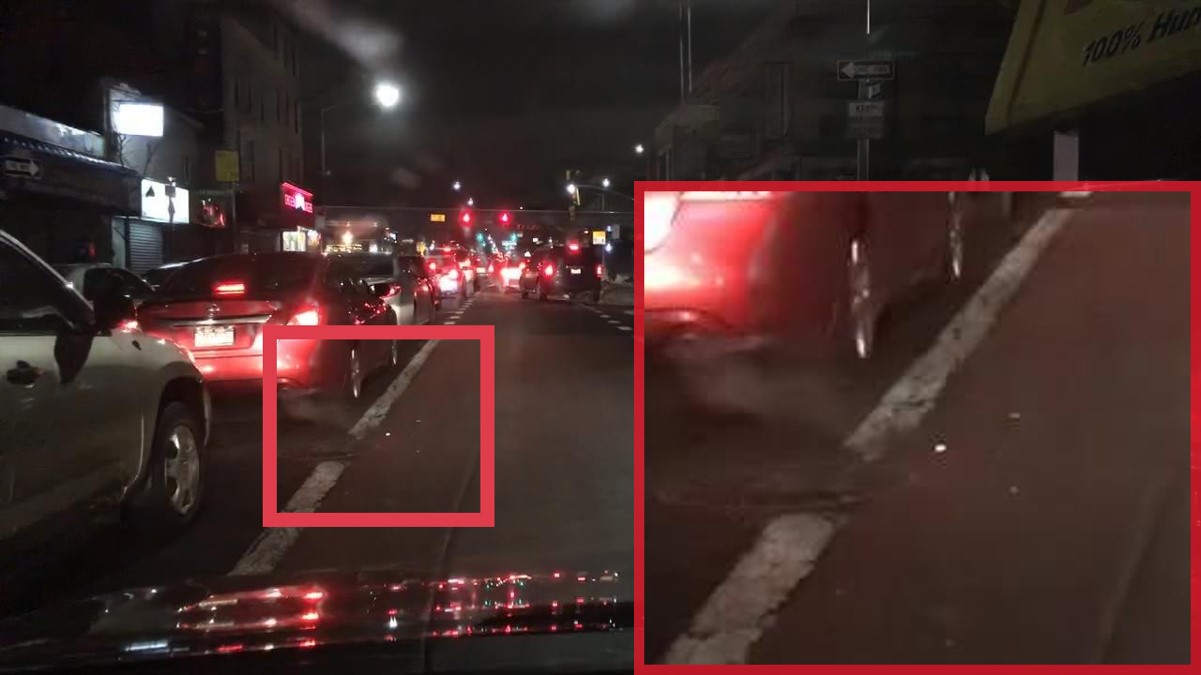}
    \end{subfigure} \\[0.5mm]

    \begin{subfigure}{0.32\linewidth}
        \centering
        \includegraphics[width=\linewidth]{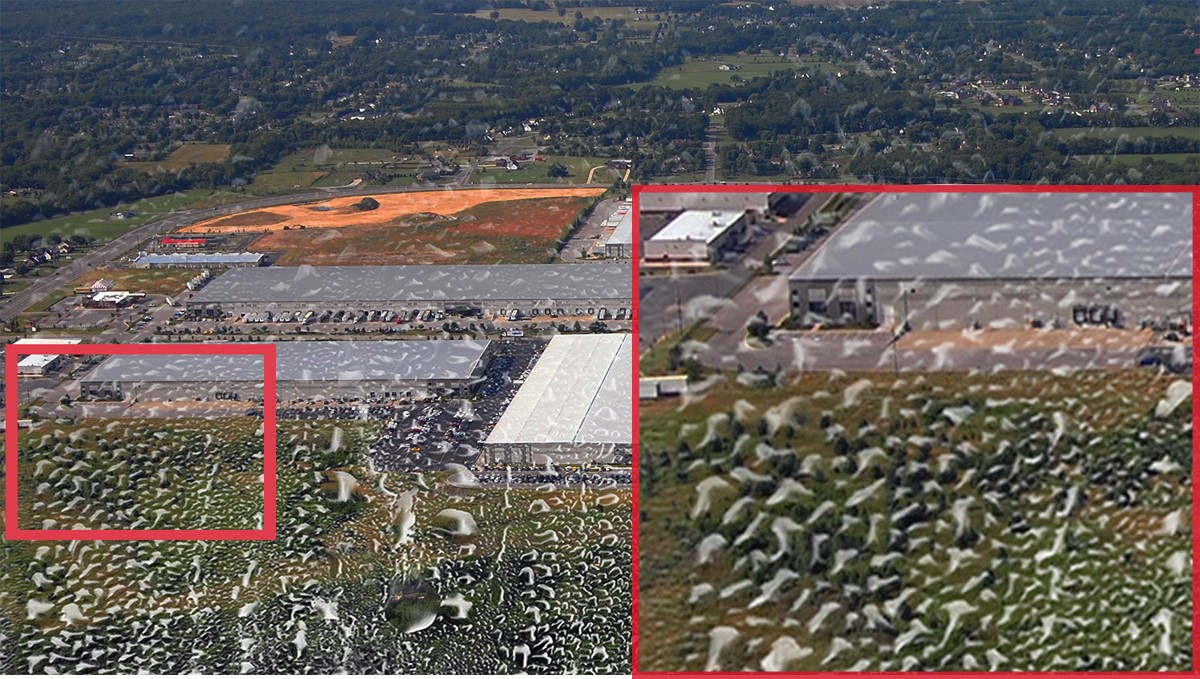}
    \end{subfigure}
    \begin{subfigure}{0.32\linewidth}
        \centering
        \includegraphics[width=\linewidth]{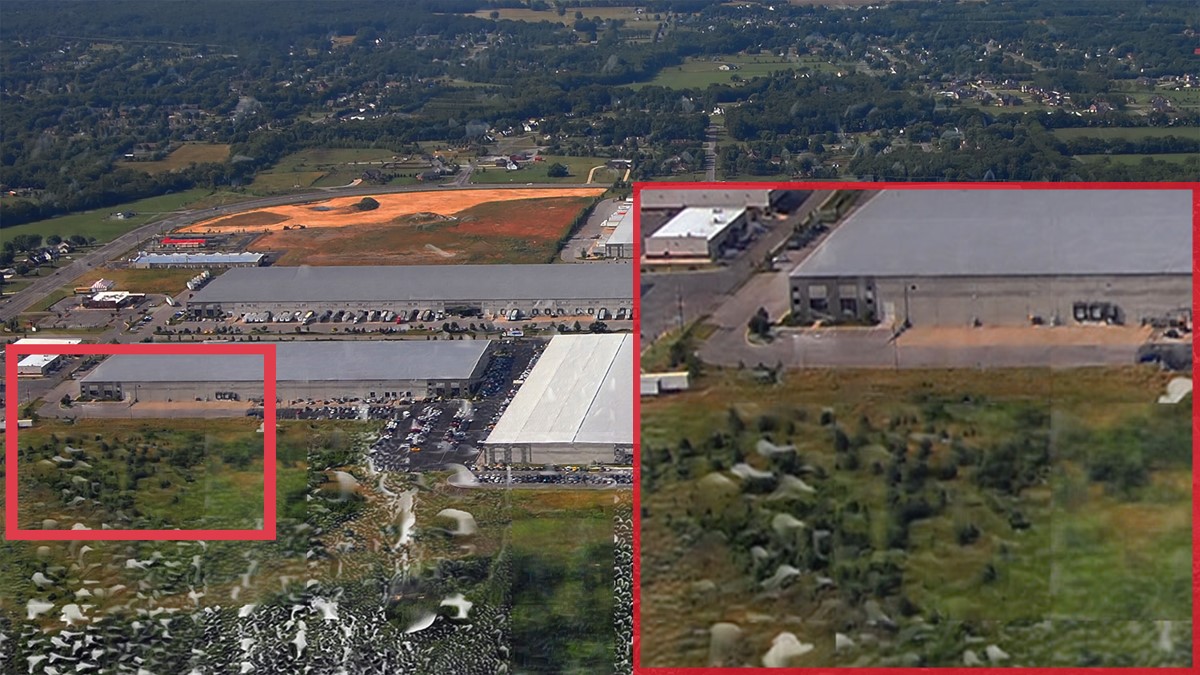}
    \end{subfigure}
    \begin{subfigure}{0.32\linewidth}
        \centering
        \includegraphics[width=\linewidth]{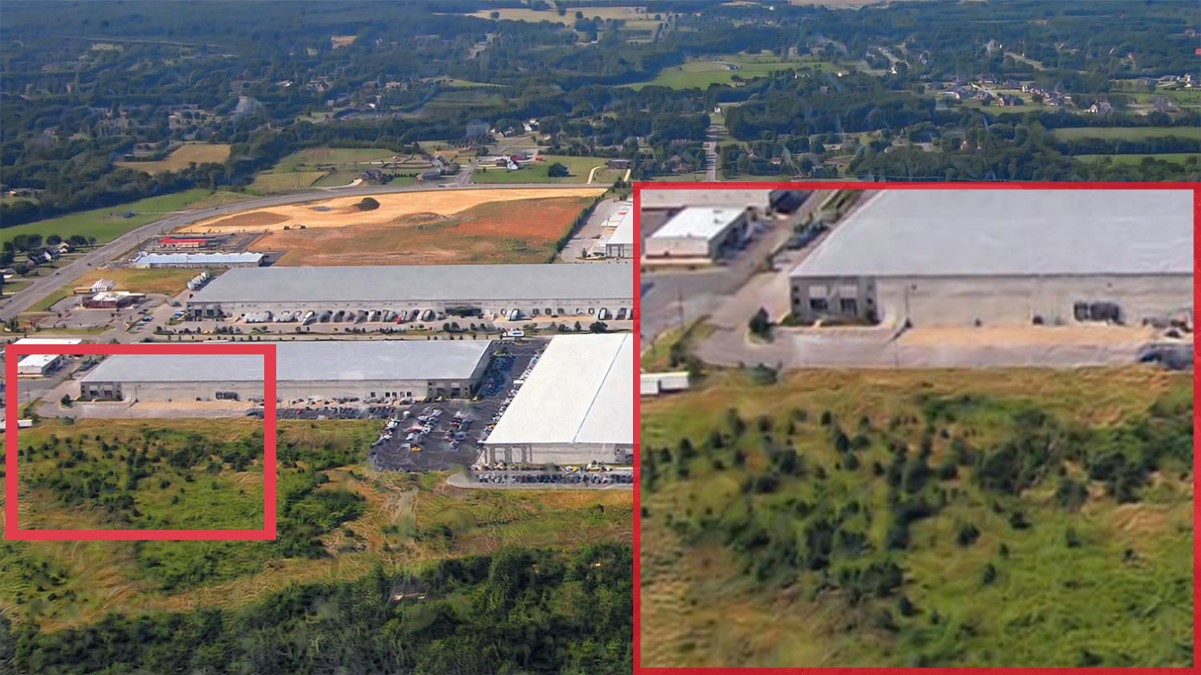}
    \end{subfigure} \\[0.5mm]

    \begin{subfigure}{0.32\linewidth}
        \centering
        \includegraphics[width=\linewidth]{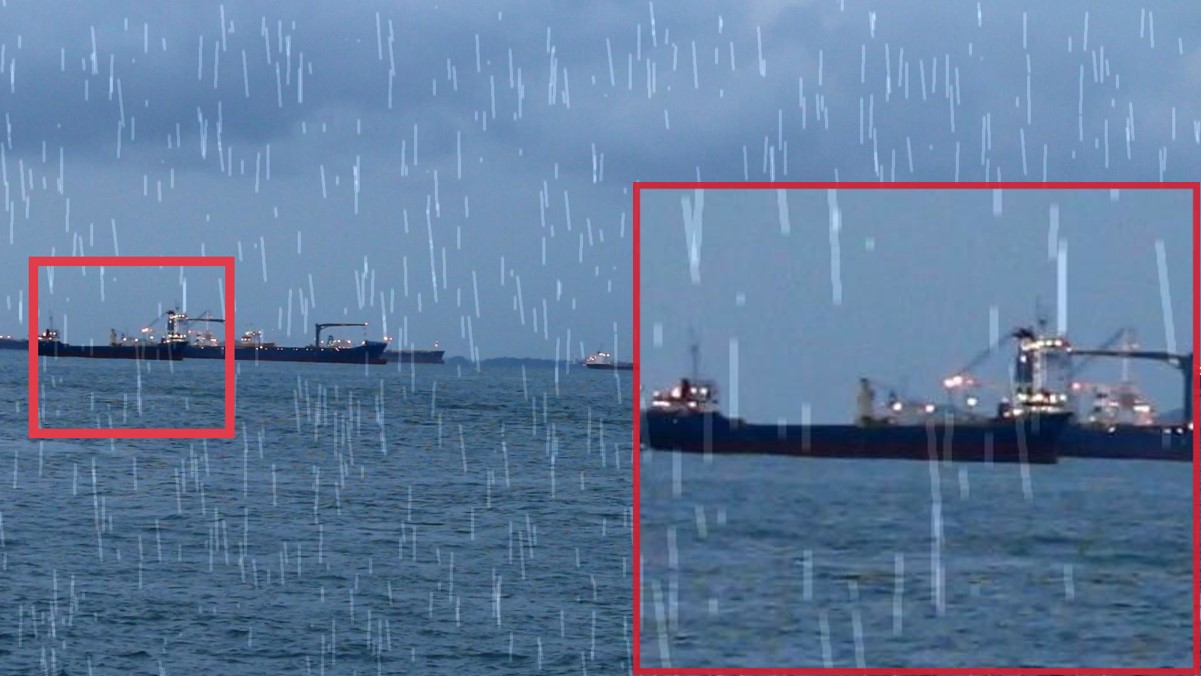}
        \subcaption*{LQ}
    \end{subfigure}
    \begin{subfigure}{0.32\linewidth}
        \centering
        \includegraphics[width=\linewidth]{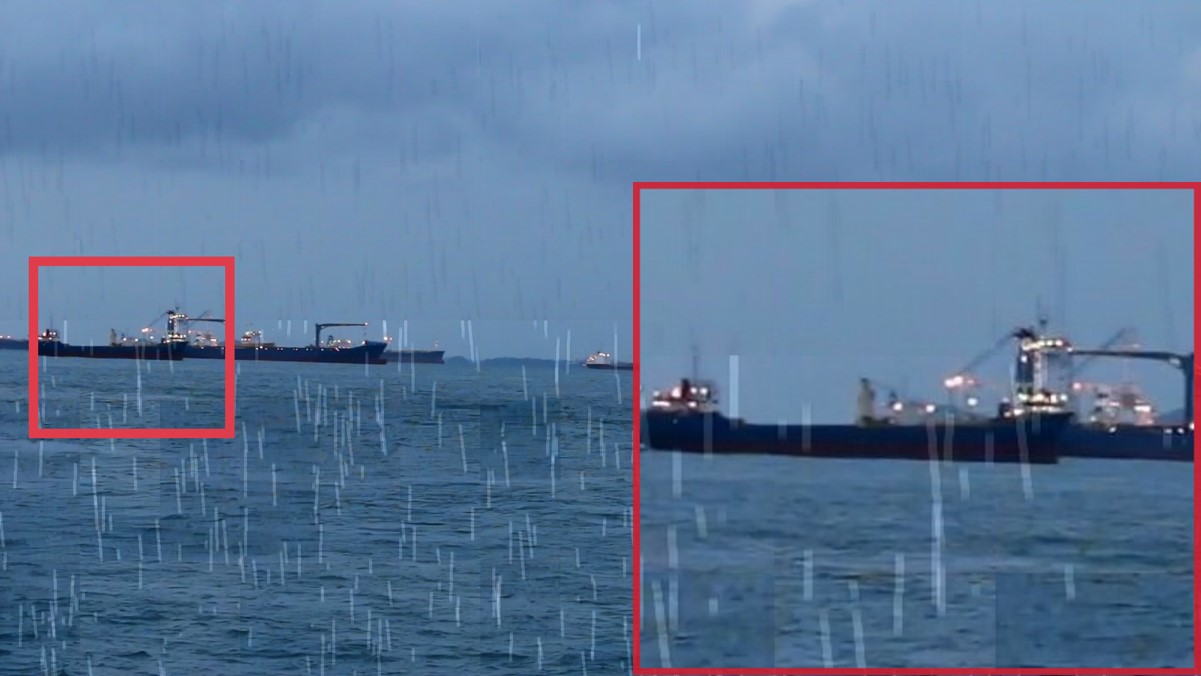}
        \subcaption*{NeRD-Rain~\cite{chen2024bidirectional}}
    \end{subfigure}
    \begin{subfigure}{0.32\linewidth}
        \centering
        \includegraphics[width=\linewidth]{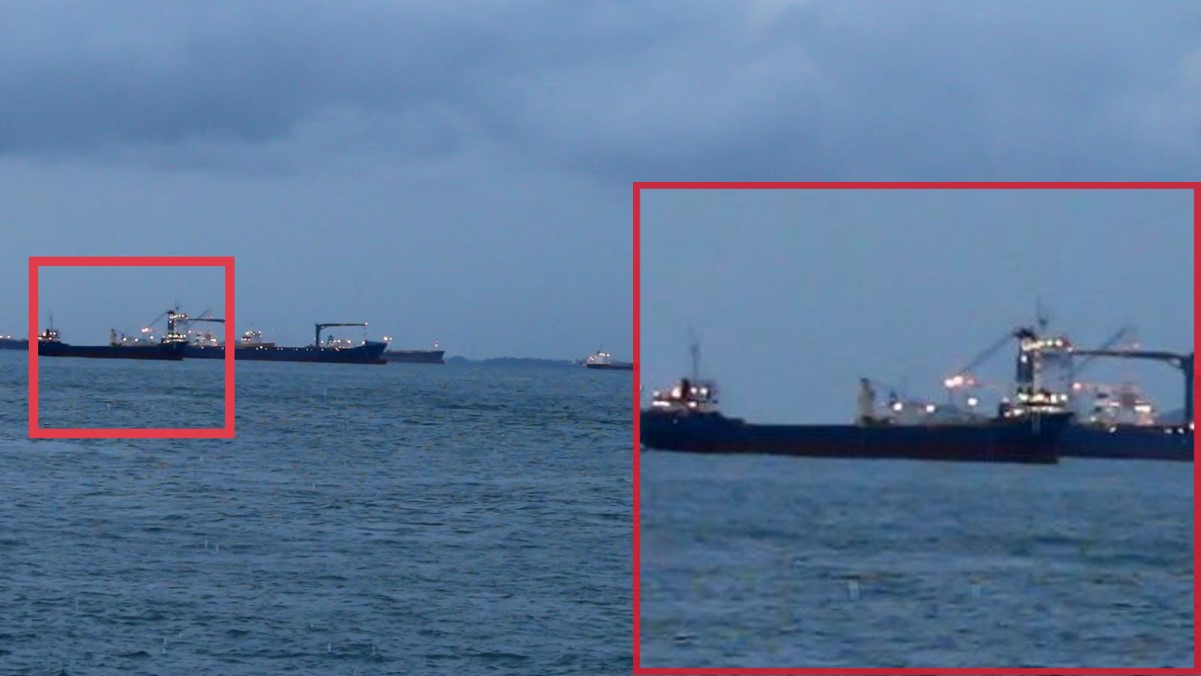}
        \subcaption*{\textbf{UniRain}}
    \end{subfigure}

    \vspace{-3mm}
    \caption{Generalization results across multiple scenarios. Our method achieves the best restoration results in driving scenes (first row), UAV scenes (second row), and maritime scenes (third row).}
    \label{fig:multi_scenarios}
    \vspace{-5mm}
\end{figure}

\vspace{-2mm}
{\flushleft\textbf{Generalization to multiple application scenarios}.}~
We further evaluate the generalization ability of our method across multiple application scenarios, including autonomous driving, unmanned aerial vehicle (UAV), and maritime scenes.
As presented in Figure~\ref{fig:multi_scenarios}, other methods struggle to preserve pixel-level structural fidelity, whereas our method produces faithful and visually pleasing results, demonstrating generalization across multiple scenarios.

\vspace{-2mm}
{\flushleft\textbf{Model complexity}.}~We evaluate the complexity of our method and state-of-the-art ones in terms of FLOPs and Params. As shown in Table~\ref{tab:flops}, UniRain achieves competitive performance with lower FLOPs and fewer parameters.

\begin{table}[htb]
\caption{Comparisons of model complexity against state-of-the-art methods.~The size of the test image is $256 \times 256$ pixels.}
\label{tab:flops}
\vspace{-3mm}
\resizebox{0.48\textwidth}{!}{
    \begin{tabular}{c|cccc}
        \toprule
        Methods       & Restormer~\cite{zamir2022restormer} & DRSformer~\cite{chen2023learning}  & NeRD-Rain~\cite{chen2024bidirectional} & UniRain (Ours)  \\ \hline
        FLOPs (G)      & 140.990   & 220.378   & 147.978   & 126.541  \\
        Params (M)     & 26.097    & 33.627    & 22.856    & 24.388  \\ \bottomrule
    \end{tabular}
    }
\vspace{-3mm}
\end{table}

\subsection{Analysis and discussion}
\label{sec:analysis}

\begin{figure}[!t]
    \centering
    \begin{subfigure}{0.48\linewidth}
        \centering
        \includegraphics[width=\linewidth]{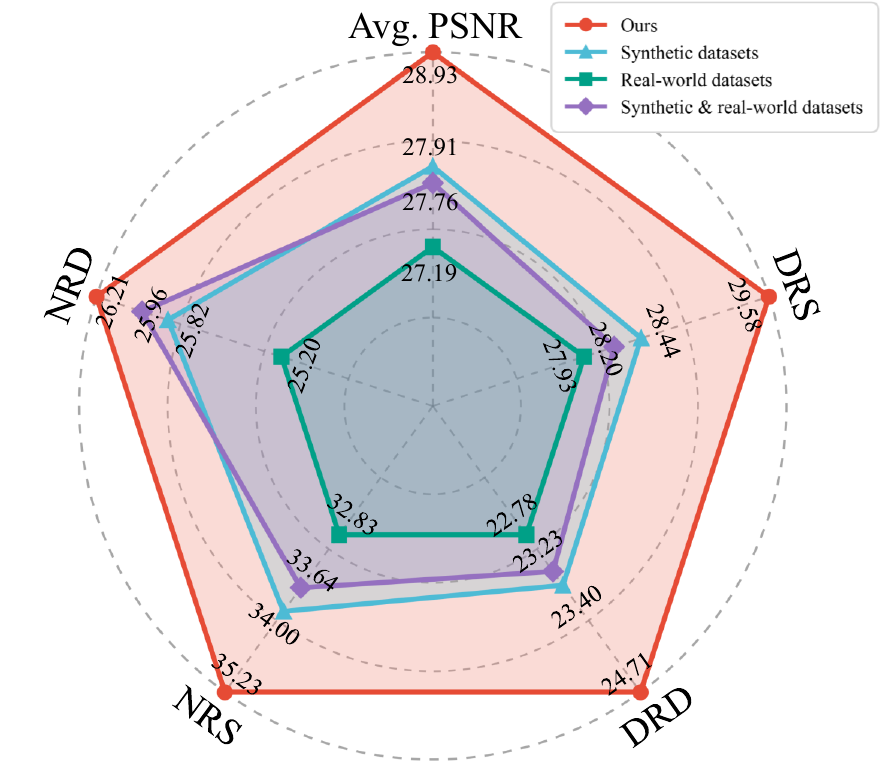}
        \caption{Different combination strategies}
        \label{fig9:image1}
    \end{subfigure}
    \begin{subfigure}{0.5\linewidth}
        \centering
        \includegraphics[width=\linewidth]{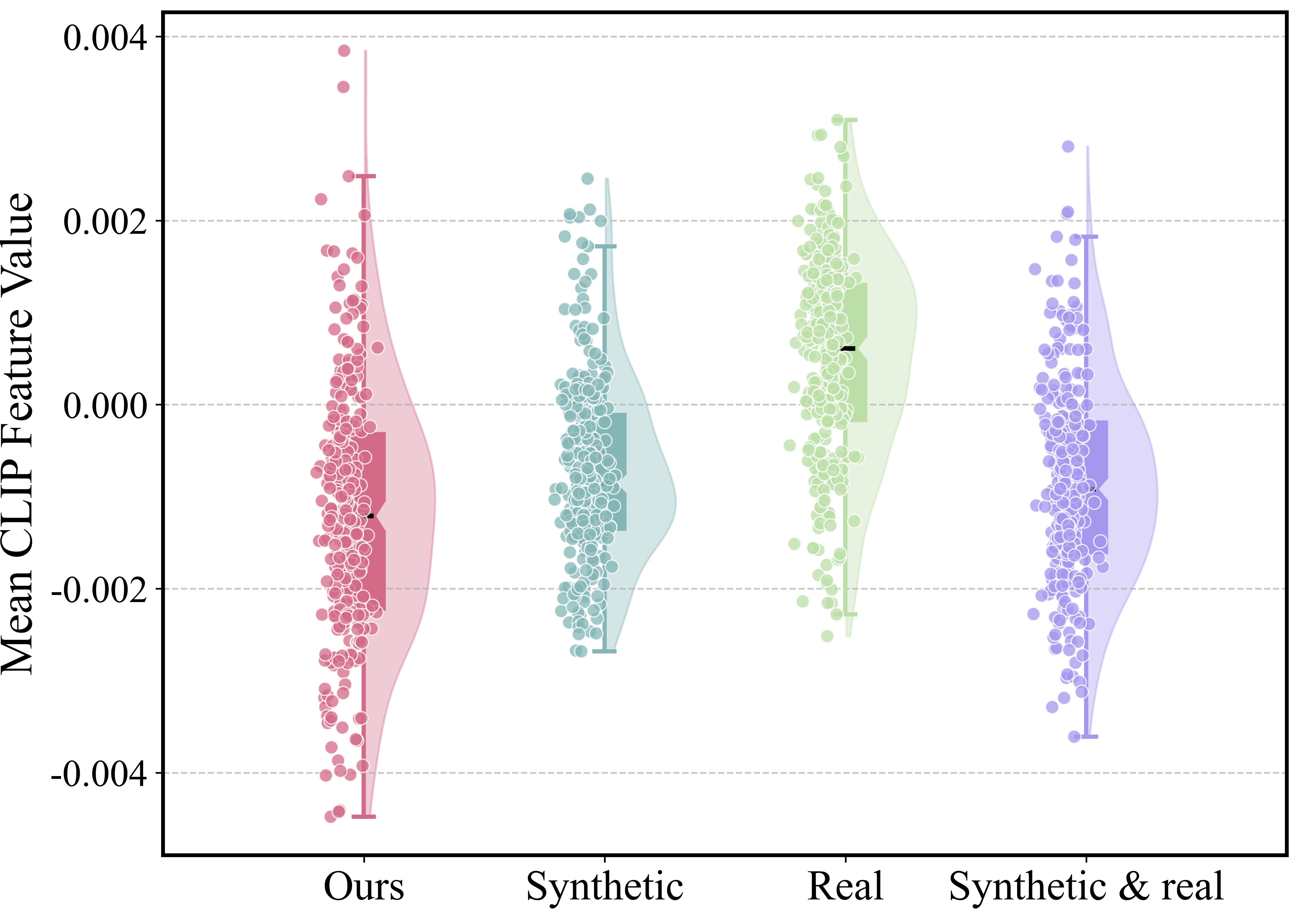}
        \caption{Feature distributions }
        \label{fig9:image2}
    \end{subfigure}
    \vspace{-3mm}
    \caption{Statistical ablation analysis of different combination strategies and their feature distributions across combinations.}
    \label{fig9}
    \vspace{-4mm}
\end{figure}

{\flushleft\textbf{Effectiveness of RAG-based dataset distillation}.}
We first analyze the impact of different mixed training datasets on the performance of the unified deraining model, including (1) synthetic dataset combinations, (2) real-world dataset combinations, and (3) synthetic $\&$ real dataset combinations.
As shown in Figure~\ref{fig9:image1}, we present the quantitative results of the same model trained on these equal-sized dataset combinations across four rain types.  
It can be observed that the model trained with our distilled dataset achieves the best overall performance, surpassing all other dataset combinations.
Figure~\ref{fig9:image2} visualizes the corresponding feature distributions, revealing that our distilled dataset covers a wider and more diverse feature space.
This richer representation contributes to stronger generalization ability and ensures more consistent deraining performance across diverse rain conditions.
Besides, we compare the averaged results between directly mixing all synthetic and real-world datasets and using our distilled dataset on the test set of Table~\ref{tab:public_dataset}.
As shown in Figure~\ref{fig:intro_b}, directly combining all datasets for training leads to suboptimal results due to inconsistent data quality across different sources.
This finding indicates that simply enlarging the dataset does not necessarily lead to better generalization. Rather, the reliability and quality of the data are the key factors that dictate model performance.

Finally, to further validate the effectiveness of our proposed RAG-based dataset distillation pipeline, we conduct a comparison with several ablated variants, including (i) using only VLMs without RAG, (ii) without retrieval stage, and (iii) without generation stage.
As shown in Table~\ref{tab:Variants}, our designed pipeline obtains higher performance as it leverages real-scene images to filter training data, ensuring that high-quality samples are distilled for model training.


\vspace{-2mm}
{\flushleft\textbf{Effectiveness of multi-objective reweighted optimization}.}
We compare our multi-objective reweighted optimization strategy with alternative schemes, including continual learning (\textit{e.g.}, PIGWM~\cite{zhou2021image}) and prompt learning (\textit{e.g.}, PromptIR~\cite{potlapalli2023promptir}). We conduct experiments on the RainDS-real-RDS~\cite{quan2021removing} dataset, with the quantitative results shown in Table~\ref{tab:Ablation_model}. It can be observed that applying our optimization strategy to the PromptIR~\cite{potlapalli2023promptir} architecture yields a 0.97 dB improvement in PSNR. Compared with continual learning, our optimization strategy improves UniRain by 1.69 dB in PSNR, highlighting its effectiveness.

To disentangle the interplay between the effectiveness of optimization strategies and that of models, we further analyze the effect of different models under the same optimization strategy. As shown in Table~\ref{tab:Ablation_model}, under the same prompt-learning optimization setting, UniRain achieves a 1.06 dB gain in PSNR over PromptIR~\cite{potlapalli2023promptir}. Moreover, under the same multi-objective reweighted optimization strategy, it surpasses PIGWM~\cite{zhou2021image} by 2.23 dB in PSNR metric.

\begin{figure}[!t]
    \centering
    \begin{subfigure}{0.49\linewidth}
        \centering
        \includegraphics[width=\linewidth]{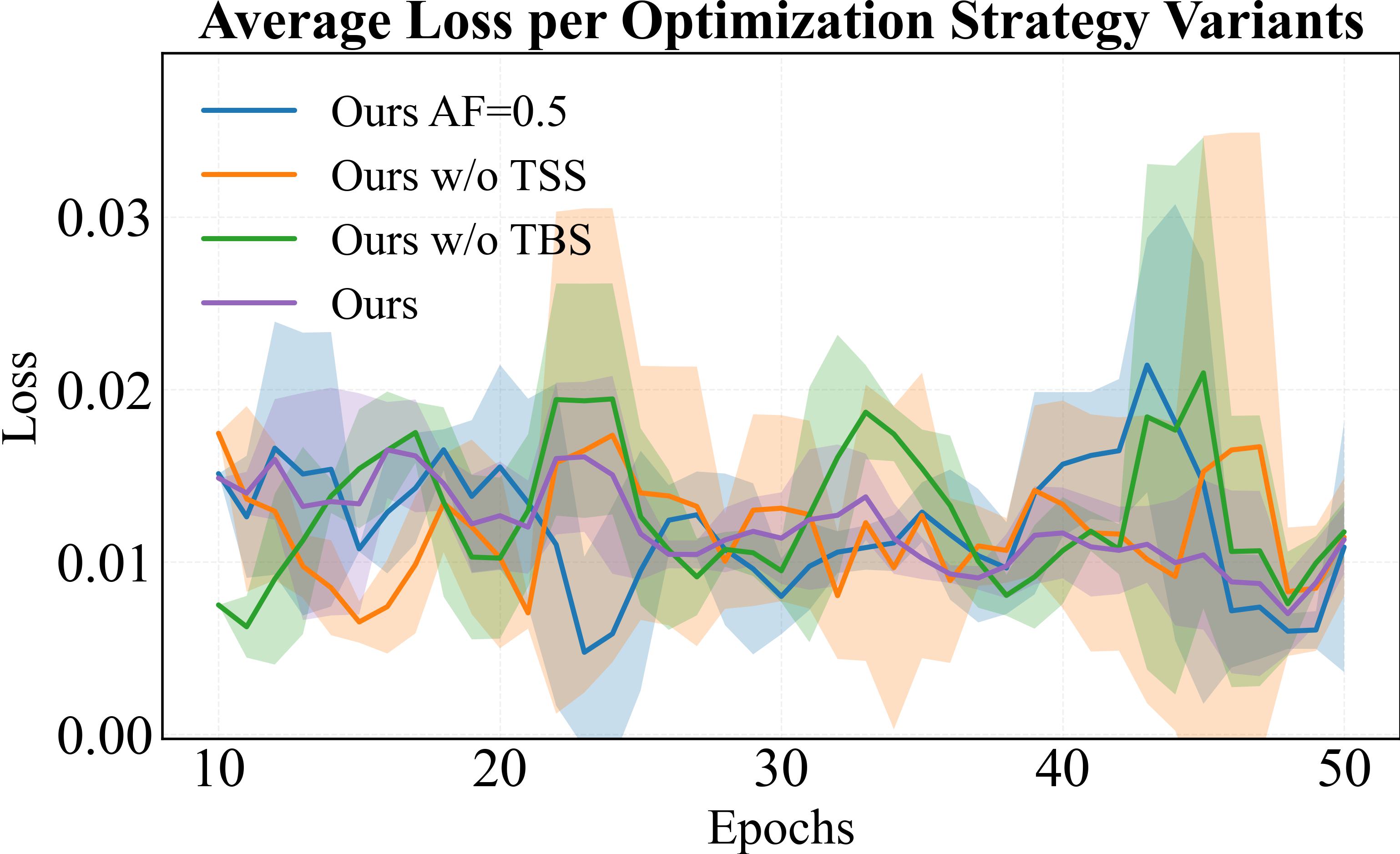}
        \caption{Average loss per variants}
        \label{fig1:optimization strategy}
    \end{subfigure}
    \begin{subfigure}{0.49\linewidth}
        \centering
        \includegraphics[width=\linewidth]{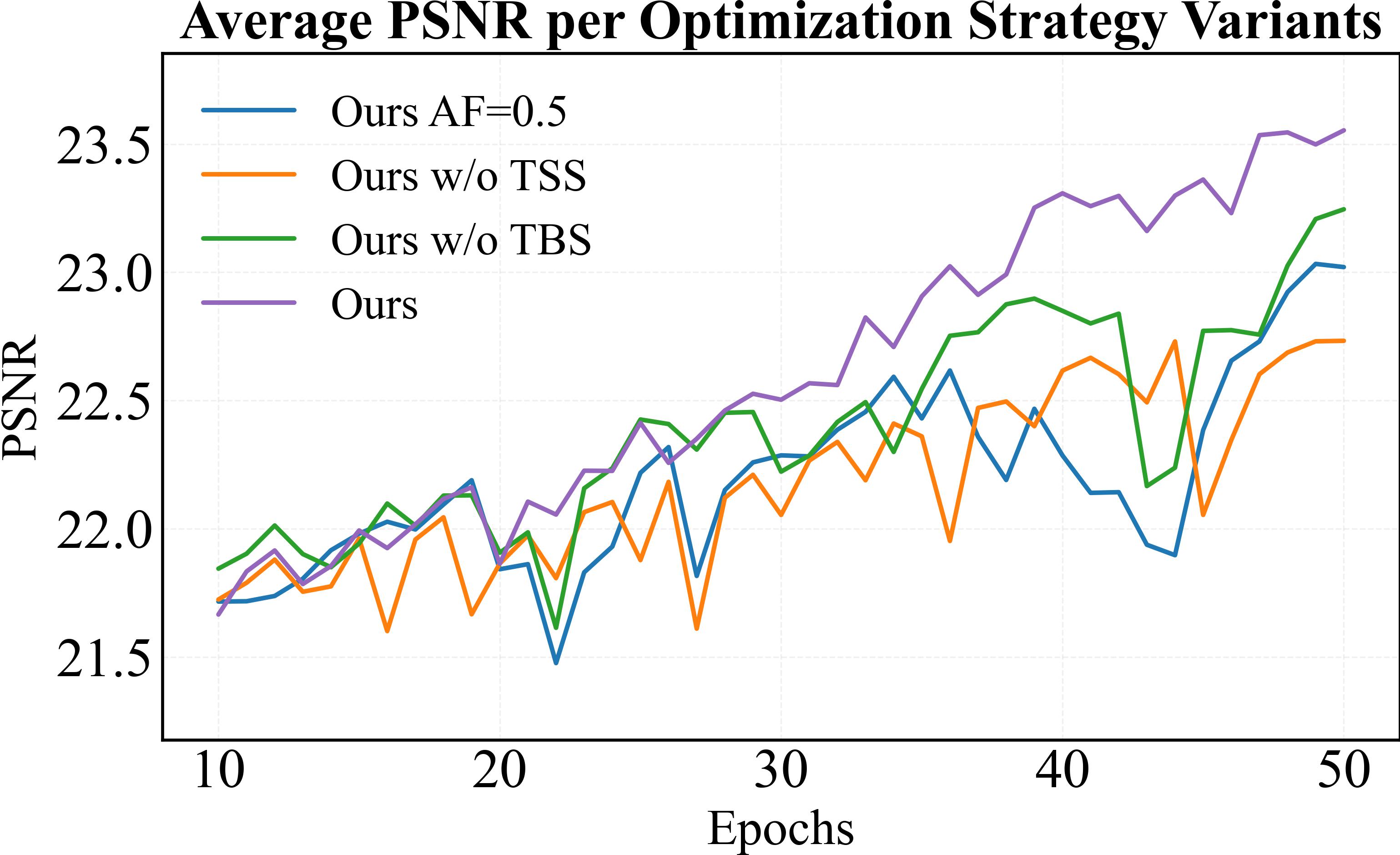}
        \caption{Average PSNR per variants}
        \label{fig2:optimization strategy}
    \end{subfigure}
    \vspace{-3mm}
    \caption{Loss and PSNR of the optimization strategy variants.}
    \label{fig:optimization strategy}
    \vspace{-4mm}
\end{figure}

Finally, we investigate the influence of different variants of the optimization strategy, including (i) a fixed adaptive factor ($\mathrm{AF}=0.5$), (ii) removing the task stability score (w/o TSS), and (iii) removing the task balance score (w/o TBS).
As illustrated in Figure~\ref{fig1:optimization strategy}, the average loss curves indicate that our full strategy achieves more stable optimization throughout training.
Meanwhile, the average PSNR results in Figure~\ref{fig2:optimization strategy} confirm that our complete formulation enables faster convergence and higher overall performance compared with the ablated variants.

\begin{table}[!t]
\caption{Ablation analysis for RAG-based dataset distillation.}
\label{tab:Variants}
\vspace{-3mm}
\resizebox{0.48\textwidth}{!}{
    \begin{tabular}{c|cccc}
        \toprule
        Variants   & only VLM   & w/o Retrieval stage   & w/o Generation stage & UniRain (Ours)  \\ \hline
        PSNR~$\uparrow$  & 27.73    & 28.39    & 28.36    & \textbf{28.93}   \\
        SSIM~$\uparrow$  & 0.8358   & 0.8453   & 0.8425   & \textbf{0.8515}    \\
        \bottomrule
    \end{tabular}
    }
\vspace{-2mm}
\end{table}

\begin{table}[!t]
\footnotesize
\centering
\caption{Ablation analysis on various models using different optimizers, including our adaptive reweighting optimization strategy.}
\vspace{-3mm}
\resizebox{0.48\textwidth}{!}{
    \begin{tabular}{c|ccc|cc}
        \toprule
        \multirow{2}{*}{\makecell{Base Models}} & \multicolumn{3}{c|}{Optimization  Strategy}        & \multicolumn{2}{c}{Metrics} \\ \cline{2-6} 
            & Continual learning  & Prompt learning  &  Ours & PSNR~$\uparrow$         & SSIM~$\uparrow$         \\ \hline
        \multirow{2}{*}{PIGWM~\cite{zhou2021image}}  & \CheckmarkBold  & \XSolidBrush    & \XSolidBrush             & 18.38   & 0.5758    \\
                                                     & \XSolidBrush    & \XSolidBrush    & \CheckmarkBold           & 18.68   & 0.5880   \\ \hline
        \multirow{2}{*}{PromptIR~\cite{potlapalli2023promptir}}  & \XSolidBrush    & \CheckmarkBold  & \XSolidBrush & 18.58   & 0.5639   \\
                                                     & \XSolidBrush    & \XSolidBrush    & \CheckmarkBold           & 19.55   & 0.5759   \\ \hline
        \multirow{3}{*}{UniRain (Ours)}                        & \CheckmarkBold  & \XSolidBrush    & \XSolidBrush   & 18.92   & 0.5657   \\
                                                     & \XSolidBrush    & \CheckmarkBold  & \XSolidBrush             & 19.64   & 0.5737   \\
                                                     & \XSolidBrush    & \XSolidBrush    & \CheckmarkBold           & \textbf{20.61}   & \textbf{0.5881}    \\ \bottomrule
    \end{tabular}
}
\label{tab:Ablation_model}
\vspace{-2mm}
\end{table}

\vspace{-2mm}
{\flushleft\textbf{Effectiveness of asymmetric MoE architecture}.}
To validate the effectiveness of asymmetric MoE architecture, we conduct ablation studies as shown in Table~\ref{Tab:MoE}.
The baseline model without any MoE modules yields the lowest performance, indicating that a plain backbone cannot effectively model complex and diverse rain degradations.
Introducing hard-MoE in both the encoder and decoder significantly enhances performance.
In contrast, using soft-MoE in both stages leads to performance degradation, suggesting that soft routing alone is insufficient to capture heterogeneous degradations.
Our design adopts soft-MoE in the encoder for exhaustive expert exploration and hard-MoE in the decoder for selective expert activation, and their integration enables the model to achieve the best overall performance.

\begin{table}[!t]
    \footnotesize
    \tabcolsep=0.1cm
    \centering
    \caption{Ablation analysis of asymmetric  MoE architecture.~Experiments are conducted on the proposed RainRAG dataset.}
    \vspace{-3mm}
    \resizebox{\linewidth}{!}{
        \begin{tabular}{c|cc|cc|cc}
            \toprule
            \multirow{2}{*}{Variants} & \multicolumn{2}{c|}{Encoder} & \multicolumn{2}{c|}{Decoder} & \multicolumn{2}{c}{Metrics} \\ \cline{2-7} 
                                     & hard-MoE& soft-MoE& hard-MoE& soft-MoE  & PSNR $\uparrow$            & SSIM $\uparrow$  \\ \hline
            Baseline  & \XSolidBrush    & \XSolidBrush   & \XSolidBrush    & \XSolidBrush   &27.54   & 0.8442  \\
            Variant 1 & \CheckmarkBold  & \XSolidBrush   & \CheckmarkBold  & \XSolidBrush   &28.39  & 0.8471  \\
            Variant 2 & \CheckmarkBold  & \XSolidBrush   & \XSolidBrush    & \CheckmarkBold &28.15   & 0.8485  \\
            Variant 3 & \XSolidBrush    & \CheckmarkBold & \XSolidBrush    & \CheckmarkBold &27.91   & 0.8465   \\
            UniRain (Ours) & \XSolidBrush & \CheckmarkBold  & \CheckmarkBold & \XSolidBrush & \textbf{28.93}  & \textbf{0.8515} \\ \bottomrule
        \end{tabular}
        }
    \vspace{-2mm}
    \label{Tab:MoE}
\end{table}

\begin{figure}[!t]
    \begin{subfigure}{0.24\linewidth}
        \centering
        \includegraphics[width=\linewidth]{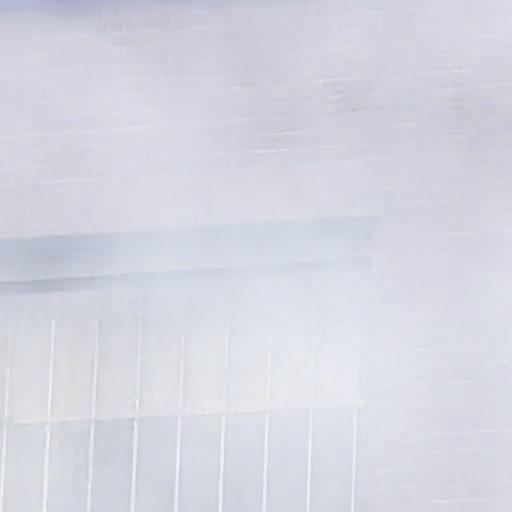}
    \end{subfigure}
    \begin{subfigure}{0.24\linewidth}
        \centering
        \includegraphics[width=\linewidth]{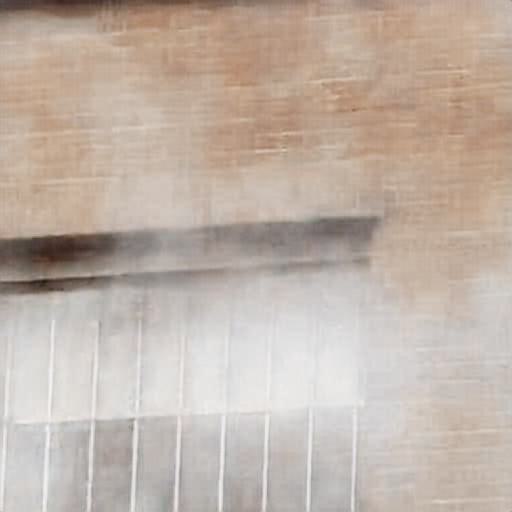}
    \end{subfigure}
    \begin{subfigure}{0.24\linewidth}
        \centering
        \includegraphics[width=\linewidth]{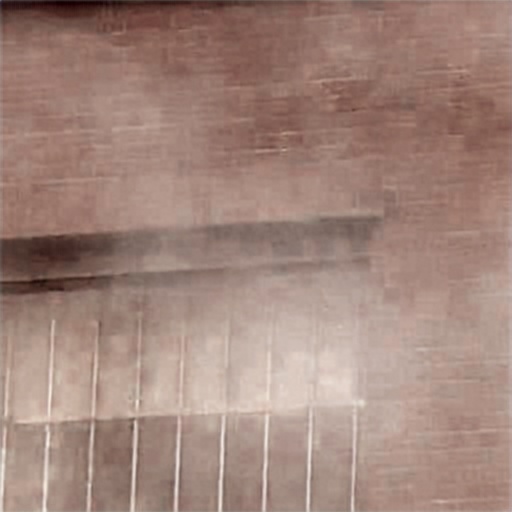}
    \end{subfigure}
    \begin{subfigure}{0.24\linewidth}
        \centering
        \includegraphics[width=\linewidth]{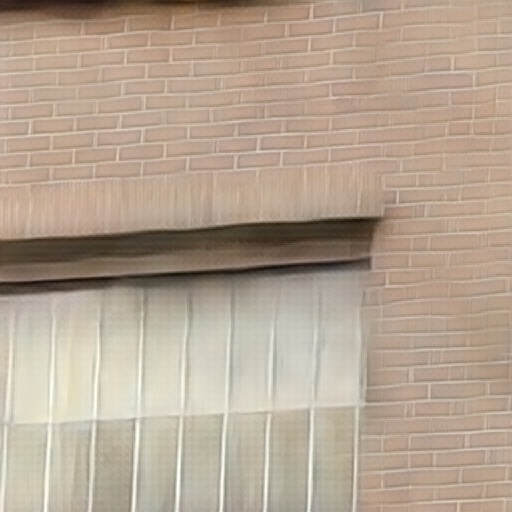}
    \end{subfigure} \\[0.5mm] 
    \begin{subfigure}{0.24\linewidth}
        \centering
        \includegraphics[width=\linewidth]{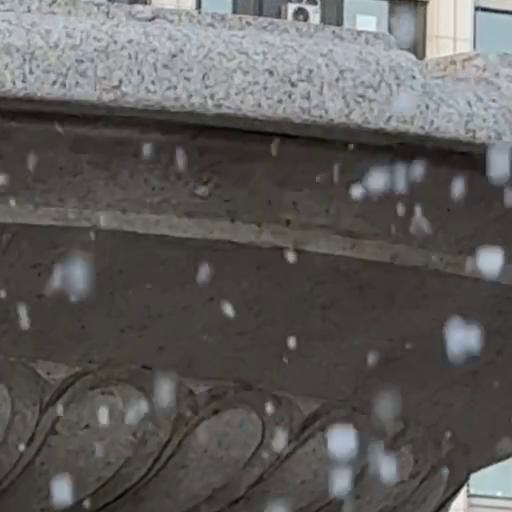}
        \subcaption*{LQ}
    \end{subfigure}
    \begin{subfigure}{0.24\linewidth}
        \centering
        \includegraphics[width=\linewidth]{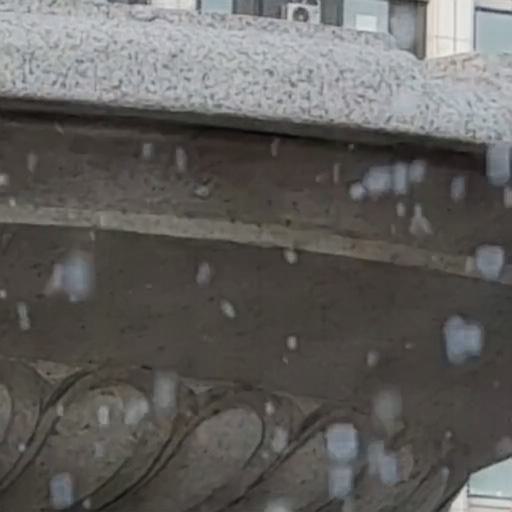}
        \subcaption*{TransWeather}
    \end{subfigure}
    \begin{subfigure}{0.24\linewidth}
        \centering
        \includegraphics[width=\linewidth]{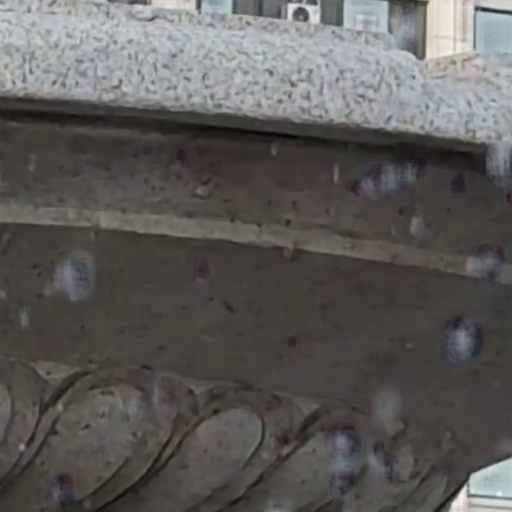}
        \subcaption*{Histoformer}
    \end{subfigure}
    \begin{subfigure}{0.24\linewidth}
        \centering
        \includegraphics[width=\linewidth]{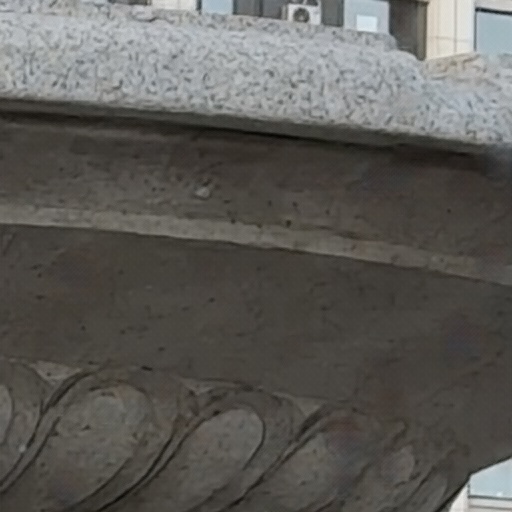}
        \subcaption*{\textbf{UniRain}}
    \end{subfigure}
    \vspace{-3mm}
    \caption{Visual comparison of all-in-one weather restoration results (\textit{e.g.}, hazy input (first row) and snowy input (second row)).
    }
    \label{fig:snowy_hazy}
    \vspace{-5mm}
\end{figure}

\vspace{-2mm}
{\flushleft\textbf{Extension to all-in-one weather restoration}.}~We extend our network to all-in-one weather restoration and compare it with adverse weather restoration method (\textit{i.e.}, TransWeather~\cite{valanarasu2022transweather}) and all-in-one image restoration approaches (\textit{i.e.}, WGWS-Net~\cite{zhu2023learning}, and Histoformer~\cite{sun2024restoring}). Table~\ref{tab:all-in-one} and Figure~\ref{fig:snowy_hazy} show that our framework achieves a clear quantitative advantage and better visual restoration results, demonstrating its effectiveness in all-in-one weather restoration.

\begin{table}[!t]
\caption{Extension to all-in-one weather restoration on the WeatherBench dataset~\cite{guan2025weatherbench}, which contains real-world multi-weather degradations (rain, snow, and haze). Average results are reported.}
\label{tab:all-in-one}
\vspace{-2mm}
\resizebox{0.48\textwidth}{!}{
    \begin{tabular}{c|ccccc}
        \toprule
        Methods          & TransWeather~\cite{valanarasu2022transweather}   & WGWS-Net~\cite{zhu2023learning}   & Histoformer~\cite{sun2024restoring}    & UniRain (Ours)  \\ \hline
        PSNR~$\uparrow$  & 24.70   & 23.89   & 24.59     & \textbf{26.01}   \\
        SSIM~$\uparrow$  & 0.7931  & 0.7811  & 0.7976    & \textbf{0.8032}    \\
        \bottomrule
    \end{tabular}
    }
\vspace{-2mm}
\end{table}

\section{Conclusion}
\label{sec:conclusion}
We have presented an effective unified framework for image deraining, called UniRain, which addresses rain streak and raindrop degradations under both daytime and nighttime conditions.
We construct an intelligent retrieval augmented generation-based dataset distillation pipeline to better improve model generalization, enabling mixed training with high-quality samples distilled from existing deraining datasets.
To achieve consistent performance and improved robustness across diverse scenarios, we integrate a simple yet effective multi-objective reweighted optimization strategy into the asymmetric union-of-experts architecture.
Extensive experiments show that our method achieves competitive performance compared with state-of-the-art models on both our proposed benchmarks and multiple public datasets.

{
    \small
    \bibliographystyle{ieeenat_fullname}
    \bibliography{main}
}


\end{document}